\documentclass[lettersize,journal]{IEEEtran}

\usepackage[utf8]{inputenc} % allow utf-8 input
\usepackage[T1]{fontenc}    % use 8-bit T1 fonts
\usepackage{url}            % simple URL typesetting
\usepackage{amsfonts}       % blackboard math symbols
\usepackage{nicefrac}       % compact symbols for 1/2, etc.
\usepackage{microtype}      % microtypography
\usepackage{xcolor}         % colors
\usepackage{multirow}
\usepackage{graphicx}
\usepackage{tikz}
\usepackage{colortbl}
\usepackage{stfloats}
\usepackage{float}
\usepackage{amssymb}
\usepackage{amsmath}
\usepackage{tabularx}
\usepackage{dsfont}
\usepackage{color}
\usepackage{amsthm}
\usepackage[ruled,linesnumbered]{algorithm2e}
\usepackage[export]{adjustbox}
\usepackage{subcaption}
\usepackage[font=footnotesize,labelfont=bf]{caption}
\usepackage{longtable}
\usepackage{tabu}
\usepackage{diagbox,booktabs}
\usepackage{multicol}
\usepackage{makecell}
\usepackage{paralist}
\usepackage{enumitem}

\definecolor{citecolor}{RGB}{65,105,225}
\usepackage[breaklinks=true,citecolor=citecolor,bookmarks=false]{hyperref}
\definecolor{royalblue}{RGB}{65,105,225} % https://www.colorabout.com/color/rgb/65,105,225/
\definecolor{lightgray}{HTML}{eeeeee}
\definecolor{tab_red}{rgb}{1,0.76,0.71}

\usepackage{dsfont}
\definecolor{dg}{rgb}{0,0.694,0.298}
\definecolor{purple}{rgb}{0.4,0.176,0.569}
\definecolor{royalblue}{RGB}{65,105,225}
\usepackage{pifont}% http://ctan.org/pkg/pifont
\makeatletter
\DeclareRobustCommand\onedot{\futurelet\@let@token\@onedot}
\def\@onedot{\ifx\@let@token.\else.\null\fi\xspace}
\def\eg{\emph{e.g}\onedot} 
\def\ie{\emph{i.e}\onedot} 
 
\def\etc{\emph{etc}\onedot} 
\def\wrt{w.r.t\onedot} 
\def\etal{\emph{et al}\onedot}

\makeatother

\definecolor{americanrose}{rgb}{1.0, 0.01, 0.24}

\title{Common Corruption Robustness of Point Cloud Detectors: Benchmark and Enhancement}
\author{Shuangzhi Li,~%~\IEEEmembership{Student Member,~IEEE,}
        Zhijie~Wang,~%~\IEEEmembership{Student Member,~IEEE,}
        Felix~Juefei-Xu,~%~\IEEEmembership{Member,~IEEE,}
        Qing~Guo*,~%~\IEEEmembership{Member,~IEEE,}
        Xingyu~Li,~%~\IEEEmembership{Member,~IEEE,}
        and Lei~Ma%,~\IEEEmembership{Member,~IEEE}
	\thanks{
	*Qing Guo is the corresponding author.
	
	Shuangzhi Li, Zhijie Wang, Xingyu Li, and Lei Ma are with the University of Alberta, AB, Canada. Zhijie Wang and Lei Ma are also with the Alberta Machine Intelligence Institute, AB, Canada. Lei Ma is also with Kyushu University, Japan. (e-mail: \{shuangzh, zhijie.wang, xingyu\}@ualberta.ca, ma.lei@acm.org)
    
    Qing Guo is with the Nanyang Technological University, Singapore. (e-mail: tsingqguo@ieee.org)
    
    Felix Juefei-Xu is with New York University, New York, NY 10012, USA. (e-mail: juefei.xu@nyu.edu)
	}% <-this % stops a space
}

\begin{document}

\maketitle

\begin{abstract}
Object detection through LiDAR-based point cloud has recently been important in autonomous driving. Although achieving high accuracy on public benchmarks, the state-of-the-art detectors may still go wrong and cause a heavy loss due to the widespread corruptions in the real world like rain, snow, sensor noise, \etc. 
Nevertheless, there is a lack of a large-scale dataset covering diverse scenes and realistic corruption types with different severities to develop practical and robust point cloud detectors, which is challenging due to the heavy collection costs.
To alleviate the challenge and start the first step for robust point cloud detection, we propose the physical-aware simulation methods to generate degraded point clouds under different real-world common corruptions.
Then, for the first attempt, we construct a benchmark based on the physical-aware common corruptions for point cloud detectors, which contains a total of 1,122,150 examples covering 7,481 scenes, 25 common corruption types, and 6 severities.
With such a novel benchmark, we conduct extensive empirical studies on 8 state-of-the-art detectors that contain 6 different detection frameworks. Thus we get several insight observations revealing the vulnerabilities of the detectors and indicating the enhancement directions.
Moreover, we further study the effectiveness of existing robustness enhancement methods based on data augmentation and data denoising. 
The benchmark can potentially be a new platform for evaluating point cloud detectors, opening a door for developing novel robustness enhancement methods.
\end{abstract}


\section{Introduction}
\label{sec:introduction}

% lidar-based object detection， 于自动驾驶场景重要
% common corruptions effect's on lidar pc and cases on detections,
% 和benchmark robustness against 的重要性
\IEEEPARstart{O}{b}ject detection via LiDAR-based point cloud \cite{guo2020deep,qian20213d}, as a crucial task in 3D computer vision, has been widely used in applications like autonomous driving \cite{badue2021self}. 
%\zhijie{add autonomous driving citation} 
Recently, the data-driven methods (\ie, deep neural networks) have significantly improved the performance of 3D point cloud detectors \cite{arnold2019survey,fernandes2021point,qian20213d}
%\zhijie{add pc detection survey citation}
on various public benchmarks, \eg, KITTI \cite{geiger2013vision}, NuScenes \cite{caesar2020nuscenes}, and Waymo \cite{sun2020scalability}.
However, the scenarios covered by these public benchmarks are usually limited. For instance, there is a lack of natural fog effects in these datasets, while fog could affect the reflection of laser beams and corrupt point cloud data with false reflections by droplets \cite{rasshofer2011influences,Bijelic_2020_STF}. Apart from the external scenarios, the internal noise of sensors can also increase the deviation and variance of ranging measurements \cite{sun2022benchmarking} and result in corrupted data and detector performance degradation. 
%Object occlusions could block beam/layer reflections \cite{xu2021behind}. 
Given that LiDAR-based point cloud detection is usually used in safety-critical applications (\eg, autonomous driving) and these external and internal corruptions could potentially affect detectors' robustness \cite{kilic2021lidar,xu2021behind,sun2022benchmarking}, it is critical to comprehensively evaluate an object detector under those corruptions before deploying it in real-world environments. 
% To this end, robustness evaluation of point cloud detectors under those common corruptions is of great importance before deployment. Hence, the robustness benchmark of point cloud detectors against common corruptions is crucially indispensable for the development of robust point cloud detectors.

% （mof）3D point benchmarks 
% CCs（mod不简写，特别是在总结和关键分析） 
% 描述问题：关键句（重复），形容限制词
% 不要大写首字母
% 

% 2D benchmark （del）(mod)
% 3D benchmark现状和局限。

% 1. First introduce robustness benchmark like extreme weather dataset. 
% 2. Limitation of directly collecting datasets. 
% 3. Success in synthesizing dataset on 2D image. 
% 4. Recent progress in ModelNet40. 
% 5. Gap between 3D object classification and detection. 
% 6. The importance of our work.

%\zhijie{complete missing citations}
There are some works constructing datasets while considering extreme weather like CADC \cite{pitropov2021canadian}, Boreas \cite{burnett2022boreas}, SeeThroughFog (STF) \cite{Bijelic_2020_STF}. 
Nevertheless, the constructed datasets only consider limited situations in the real world due to the heavy collection costs, which are far from a comprehensive evaluation. For instance, Boreas only covers 4 rainy scenes and 5 snowy scenes. STF only contains foggy point clouds at severity levels of ``dense'' and ``light''.
Hence, there is an increasing demand for extending existing benchmarks to conduct a comprehensive evaluation through covering diverse corruptions in the real world. 
A straightforward way is to synthesize the corrupted point clouds given the success of similar solutions in the image-based tasks \cite{hendrycks2019benchmarking,barbu2019objectnet} and 3D object recognition \cite{sun2022benchmarking, ren2022benchmarking}. 
However, there is no accessible dataset for the robustness evaluation of point cloud detectors. 
Note that, the robustness datasets (\eg, Modelnet40-C \cite{sun2022benchmarking}) for 3D object recognition cannot be used to evaluate the point cloud detectors, directly: (1) the example in the recognition dataset only contains the points of an object and cannot be adopted for object detection task that aims to localize and classify objects in 3D scene. (2) The latest Modelnet-C \cite{ren2022benchmarking} and Modelnet40-C \cite{sun2022benchmarking}) only consider 7 corruptions and 15 corruptions, respectively, which is still limited for a comprehensive evaluation in safety-critical environments such as autonomous driving.
%

% challenge 1: 包含全面corruption的dataset
The main challenge for building a dataset for the robustness evaluation of point cloud detection stems from the huge amount of diverse corruption types with different physical imaging principles.
For example, flawed sensors and different object characteristics could lead to noise-like corruptions and affect spherical and Cartesian coordinates of points, respectively. 
Different weathers like rain and fog might lead to false reflections. 
These corruptions have different imaging principles and need careful designs of the respective simulation methods. 

\begin{table*}[t]
\centering
\caption{{Summary of datasets used for LiDAR-based point cloud object detection}}
\label{tab: summary_dataset}
\resizebox{\textwidth}{!}{%
\begin{tabular}{@{}lccccclcc@{}}
\toprule
\textbf{Dataset} & \textbf{Year} & \textbf{Real/Simulated} & \textbf{Frames} & \textbf{BBoxes} & \textbf{Classes} & \multicolumn{1}{c}{\textbf{Corruptions}} & \textbf{\begin{tabular}[c]{@{}c@{}}Corruption\\ Severities\end{tabular}} & \textbf{\begin{tabular}[c]{@{}c@{}}Robustness\\ Metric\end{tabular}} \\ \midrule
\textbf{KITTI} \cite{geiger2013vision} & 2012 & real & 15K & 200K & 8 & cutout, noise & 2 & - \\ \cmidrule(r){7-7} 
\textbf{NuScenes} \cite{caesar2020nuscenes} & 2019 & real & 400K & 1.4M & 23 & rain, sun, clouds, cutout, various vehicle types, noise & 2 & - \\ \cmidrule(r){7-7} 
\textbf{Waymo} \cite{sun2020scalability} & 2019 & real & 200K & 12M & 4 & rain, fog, cutout, dust, various vehicle types, noise & 2 & - \\ \cmidrule(r){7-7} 
\textbf{Boreas} \cite{burnett2022boreas} & 2022 & real & 7.1K & 320K & 4 & snow, rain, sun, clouds, cutout, noise & 2 & - \\ \cmidrule(r){7-7} 
\textbf{STF} \cite{Bijelic_2020_STF} & 2020 & real & 13.5K & 100K & 4 & fog, rain, snow, cutout, noise & 3 & - \\ \cmidrule(r){7-7}
\textbf{CADC} \cite{pitropov2021canadian} & 2020 & real & 7K & 334K & 10 & snow, bright light, cutout, noise & 5 & - \\ \cmidrule(r){7-7} 

\textbf{ModelNet40-C} \cite{sun2022benchmarking} & 2022 & real+simulated & 185K & - & 40 & \begin{tabular}[c]{@{}l@{}}occlusion, LiDAR, local\_density\_inc/dec, cutout,\\uniform, Gaussian, impulse, upsampling, background,\\ rotation, shear, FFD, RBF, inv\_RBF\end{tabular} & 6 & \checkmark \\ \cmidrule(r){7-7} 

\textbf{ModelNet-C} \cite{ren2022benchmarking} & 2022 & real+simulated & 185K & - & 40 & \begin{tabular}[c]{@{}l@{}} scale, rotate, jitter, drop\_global/local, add\_global/local \end{tabular} & 6 & \checkmark \\ \cmidrule(r){7-7} 

\textbf{Argoverse} \cite{chang2019argoverse} & 2019 & real & 468K & 993K & 15 & rain, cutout, dust, noise & 2 & - \\ \cmidrule(r){7-7} 
\textbf{Lyft Level 5} \cite{Woven@2019} & 2020 & real & 30K & 1.3M & 9 & rain, cutout, noise & 2 & - \\ \midrule 
\textbf{Ours} & 2022 & real+simulated & \cellcolor{tab_red}1.1M & 15M & 8 & \begin{tabular}[c]{@{}l@{}}\textbf{Scene}: rain, snow, fog, uniform\_rad, gaussian\_rad,\\ impulse\_rad, upsample, background, cutout, beam\_del,\\ local\_dec/inc, layer\_del; \textbf{Object}: uniform, gaussian,\\ impulse, upsample, cutout, local\_dec/inc, shear, scale,\\ rotation, FFD, translation\end{tabular} & 6 & \checkmark \\ \bottomrule
\end{tabular}%
}
\end{table*}

In this work, for the first attempt, we construct a benchmark to evaluate the robustness of point cloud object detectors based on LiDAR under diverse common corruptions and discuss the effectiveness of existing robustness enhancement methods.
Regarding the benchmark construction, we first design physical-aware simulation methods for $25$ corruptions according to their physical models, respectively. Then, we borrow 7,481 raw 3D scenes (\ie, clean point clouds) from \cite{geiger2013vision} and build large-scale corrupted datasets by adding $25$ corruptions with $6$ different severity levels to each clean point cloud. Finally, we obtain a total of 1,122,150 examples covering 7,481 scenes, 25 common corruption types, and 6 severity levels. 
% {Compared with other benchmarks shown in Table \ref{tab: summary_dataset}, our benchmark covers more types of corruptions and synthesized more data, comprehensively benchmarking the robustness of point cloud detection against common corruptions.}
{Compared with real-world data benchmark (see Table~\ref{tab: summary_dataset}), the proposed benchmark synthesized more examples for benchmarking robustness. Compared with other synthesized benchmark (see Table~\ref{tab: summary_dataset}), our benchmark provides more types of corruption patterns to specifically support benchmarking object detection.}
%
% While synthesizing real-world corruptions could be challenging, we leverage the physical-aware simulation methods to generate degraded
% point clouds under different real-world common corruptions. 
%
Note that, we conduct extensive experiments to quantitatively validate the effectiveness of simulation methods by evaluating the naturalness of synthesized data.
% %
% Based on these corruption synthesis methods, we then build a physical-aware benchmark for point cloud detectors, which 

With such a novel benchmark, we investigate the robustness of current point cloud detectors by conducting extensive empirical studies on 8 existing detectors, covering 3 different representations and 2 different proposal architectures. In particular, we study the following four research questions to identify the challenges and potential opportunities for building robust point cloud detectors:

\begin{itemize}[leftmargin=*]
    \item \textbf{How do the common corruption patterns affect the point cloud detector’s performance?} Given overall common corruptions, an accuracy drop of $11.01\%$ (on average) on all detectors anticipates a noticeable accuracy drop of detectors against diverse corruption patterns.
    \item \textbf{How does the design of a point cloud detector affect its robustness against corruption patterns?} Compared with two-stage detectors, one-stage detectors perform more robust against a majority of corruptions. {Compared with point-based detectors, voxel-involving detectors perform more robust against the most of corruptions.}
    \item \textbf{What kind of detection bugs exist in point cloud detectors against common corruption patterns?} Followed by the decrease in the rate of true detection, common corruptions widely trigger a number of false detections on all point cloud detectors.
    \item \textbf{How do the robustness enhancement techniques improve point cloud detectors against common corruption patterns?} Even with the help of data augmentation and denoising, common corruptions still cause a severe accuracy drop of over $10\%$ on detection.
\end{itemize}

% In the end, we get several insight observations revealing the vulnerabilities of the detectors and illustrate the limited effectiveness of existing robustness enhancement methods based on data augmentation and active data denoising. 
% our contributions.
In summary, this work makes the following contributions:
\begin{itemize}[leftmargin=*]
    \item We design physical-aware simulation methods covering 25 common corruptions related to natural weather, noise disturbance, density change, and object transformations at the object and scene level.
    \item We create the first robustness benchmark of point cloud detection against common corruptions.
    \item Based on the benchmark, we conduct extensive empirical studies to evaluate the robustness of 8 existing detectors to reveal the vulnerabilities of the detectors under common corruptions. 
    \item We study the existing data augmentation (DA) method and denoising method's performance on robustness enhancement for point cloud detection and further discuss their limitations.
\end{itemize}

\begin{table*}[t]
\caption{Taxonomy of collected common corruption patterns}
\label{tab: corruption details}
\centering
\resizebox{\textwidth}{!}{
\scriptsize
\begin{tabular}{@{}c|c|l||c|c|l@{}}
\toprule
\multicolumn{3}{c||}{\textbf{Scene-level}} & \multicolumn{3}{c}{\textbf{Object-level}} \\ \midrule
\textbf{\begin{tabular}[c]{@{}c@{}}Corruption\\Category\end{tabular}} & \textbf{Corruption} & \multicolumn{1}{c||}{\textbf{Potential Reasons}} & \textbf{\begin{tabular}[c]{@{}c@{}}Corruption\\Category\end{tabular}} & \textbf{Corruption} & \multicolumn{1}{c}{\textbf{Potential Reasons}} \\ \midrule
\multirow{3}{*}{\textbf{Weather}} & \textit{rain} & \multirow{3}{*}{\textbf{Environment}: natural weather \cite{rasshofer2011influences};} & \multirow{4}{*}{\textbf{Noise}} & \textit{uniform} & \multirow{4}{*}{\begin{tabular}[l]{@{}l@{}} \textbf{Object surface}: coarse surface \cite{bolkas2018effect} \\and dark-color cover \cite{bolkas2018effect};\end{tabular}} \\ \cmidrule(lr){2-2} \cmidrule(lr){5-5}
 & \textit{snow} &  &  & \textit{gaussian} &  \\ \cmidrule(lr){2-2} \cmidrule(lr){5-5}
 & \textit{fog} &  &  & \textit{impulse} &  \\ \cmidrule(r){1-3} \cmidrule(lr){5-5}
\multirow{5}{*}{\textbf{Noise}} & \textit{uniform\_rad} & \multirow{4}{*}{\begin{tabular}[l]{@{}l@{}}\textbf{Environment}: strong illumination \cite{villa2021spads};\\ \textbf{Sensor}: low ranging accuracy \cite{Texas2016lidar} and\\ sensor vibration \cite{ma2012analysis,wang2021simultaneous};\end{tabular}} &  & \textit{upsample} &  \\ \cmidrule(lr){2-2} \cmidrule(l){4-6} 
 & \textit{gaussian\_rad} &  & \multirow{3}{*}{\textbf{Density}} & \textit{cutout} & \multirow{3}{*}{ \begin{tabular}[l]{@{}l@{}} \textbf{Object surface}: object or self-\\occlusions \cite{xu2021behind}, dark-color cover \cite{bolkas2018effect}\\ and transparent components;\end{tabular}} \\ \cmidrule(lr){2-2} \cmidrule(lr){5-5}
 & \textit{impulse\_rad} &  &  & \textit{local\_dec} &  \\ \cmidrule(lr){2-2} \cmidrule(lr){5-5}
 & \textit{upsample} &  &  & \textit{local\_inc} &  \\ \cmidrule(lr){2-3} \cmidrule(l){4-6} 
 & \textit{background} & \textbf{Environment}: floating particles \cite{mona2012lidar}; & \multirow{5}{*}{\textbf{Transformation}} & \textit{translation} & \multirow{2}{*}{\begin{tabular}[l]{@{}l@{}}\textbf{Object}: different locations and\\ heading directions \cite{morin2021simulated};\end{tabular}} \\ \cmidrule(r){1-3} \cmidrule(lr){5-5}
\multirow{5}{*}{\textbf{Density}} & \textit{cutout} & \multirow{5}{*}{\begin{tabular}[l]{@{}l@{}}\textbf{Sensor}: different scanning layers, object\\occlusion \cite{xu2021behind}, and randomly laser beam\\ \cite{xu2021behind} or layer (rotary laser) malfunction;\end{tabular}} &  & \textit{rotation} &  \\ \cmidrule(lr){2-2} \cmidrule(l){5-6} 
 & \textit{local\_dec} &  &  & \textit{shear} & \multirow{3}{*}{\begin{tabular}[l]{@{}l@{}} \textbf{Object deformation}: bending or\\ moving pedestrians \cite{wong2020efficient}, different\\styles of vehicles \cite{wang2020train}.\end{tabular}} \\ \cmidrule(lr){2-2} \cmidrule(lr){5-5}
 & \textit{local\_inc} &  &  & \textit{FFD} &  \\ \cmidrule(lr){2-2} \cmidrule(lr){5-5}
 & \textit{beam\_del} &  &  & \textit{scale} &  \\ \cmidrule(lr){2-2} \cmidrule(l){4-6} 
 & \textit{layer\_del} &  & \multicolumn{1}{l}{} & \multicolumn{1}{l}{} & \multicolumn{1}{l}{} \\ \bottomrule
\end{tabular}%
}
\end{table*}

\section{Related Work}
\subsection{LiDAR Perception}
\label{sec: lidar_perception}
{
LiDAR perception is sensitive to both internal and external factors that could result in different corruptions. Adversarial weather \cite{rasshofer2011influences} (\eg, snow, rain, and fog) can dim or even block transmissions of lasers by dense liquid or solid droplets. 
Regarding noise characteristics of point clouds, strong illumination \cite{villa2021spads} affects the signal transmission by lowering Signal-to-Noise Ratio (SNR), increasing the noise level of LiDAR ranging \cite{Texas2016lidar}. Besides, the intrinsically inaccurately ranging and the sensor vibration \cite{ma2012analysis,wang2021simultaneous} potentially trigger noisy observations during LiDAR scanning. Environmental floating particles (\eg, dust \cite{mona2012lidar}) could perturb point cloud with the background noise. 
Density distribution of LiDAR-based point clouds can also easily affect autonomous driving. For instance, common object-object occlusions block LiDAR scanning on objects in the scene \cite{xu2021behind}. Besides, the dark-color cover and rough surface \cite{bolkas2018effect} could affect LiDAR's reflection and thus reduce local point density when sensing such objects. Moreover, the malfunction of (fixed or rotary) lasers \cite{xu2021behind} globally loses points or layers of points in point clouds.
For 3D tasks, various shapes \cite{wong2020efficient,wang2020train}, locations and poses \cite{morin2021simulated} of objects can also influence the context perception in the scene. }

{
Apart from these natural corruptions, LiDAR perception is also sensitive to adversarial attack. Adversarial attacks \cite{xiang2019generating} pose significant security issues and vulnerability on 3D point cloud tasks (\eg, classification \cite{liu2019extending}, detection \cite{abdelfattah2021adversarial}, and segmentation \cite{zhu2021adversarial}). 
}

% mod： 去掉（3D） （去掉self-training）
% pc，huati+ 加粗 =》 加粗
% 换掉j，成（n,m）

\subsection{Point Cloud Detectors}
% description
% representation, mof（del pros and cons）
% proposal structures, pros and cons

%Point cloud detection serves as a non-trivial task of locating objects of interest in the format of 3D bounding boxes (BBoxes). 
Based on the different representations acquired from point clouds, point cloud detectors can be categorized into \textbf{2D-view-based} detectors (\eg, VeloFCN \cite{li2016vehicle} and PIXOR \cite{yang2018pixor}), \textbf{voxel-based detectors} (\eg, SECOND \cite{yan2018second} and VoTr \cite{mao2021voxel}), \textbf{point-based} detectors (\eg, PointRCNN \cite{shi2019pointrcnn} and 3D-SSD \cite{yang20203dssd}), and \textbf{point-voxel-based} detectors (\eg, PVRCNN \cite{shi2020pv} and SA-SSD \cite{he2020structure}). On the other hand, based on the different proposal architectures, point cloud detectors can also be divided into \textbf{one-stage} detectors (\eg, 3D-SSD \cite{yang20203dssd} and SA-SSD \cite{he2020structure}) and \textbf{two-stage} detectors (\eg, PointRCNN \cite{shi2019pointrcnn} and PVRCNN \cite{shi2020pv}). In this paper, we select 8 representative methods covering all these categories. 

\subsection{Robustness Benchmarks against Common Corruptions}
% 2D benchmark
% 3D benchmark and 问题
Several attempts have been made to benchmark robustness for different data domains. Based on ImageNet \cite{deng2009imagenet}, ImageNet-C simulates real-world corruptions to test image classifiers' robustness. ObjectNet \cite{barbu2019objectnet} illustrates the performance degradation of 2D recognition models considering object backgrounds, rotations, and imaging viewpoints. Inspired by 2D works, several benchmarks were built for 3D tasks. Modelnet40-C \cite{sun2022benchmarking} corrupts ModelNet40 \cite{wu20153d} with 15 simulated common corruptions affecting point clouds’ noise, density, and transformations, to evaluate the robustness of point cloud recognition. Targeting 7 fundamental corruptions (\ie, ``Jitter'', ``Drop Global/Local'', ``Add Global/Local'', ``Scale'', and ``Rotate''), ModelNet-C reveals the vulnerability of different components of 9 existing point cloud classifiers. Regarding point cloud detection, NuScenes, Waymo, and STF collect LiDAR scans under adversarial rainy, snowy, and foggy conditions, where the accuracy of 3D detectors is tested \cite{caesar2020nuscenes, Bijelic_2020_STF, sun2020scalability}. However, to the best of our knowledge, a lack of benchmark of point cloud detection's robustness comprehensively against various common corruptions is still remaining. 

\subsection{Robustness Enhancement for Point Cloud Detection}
% 介绍 3D classification的 enhancing 方法
% 介绍 detection的Enhancement的几种方式和具体方法：augmentation， self training， denoising
% 问题
Recently, improving the robustness of point cloud detection has also received significant concerns. Zhang \etal propose PointCutMix \cite{zhang2021pointcutmix} as a single way to generate new training data by replacing the points in one sample with their optimal assigned pairs in another sample. Lee \etal \cite{lee2021regularization} propose a rigid subset mix (RSMix) augmentation to get a virtual mixed sample by replacing part of the sample with shape-preserved subsets from another sample. Specifically for 3D object detection, there are several ways to improve detectors' robustness. Choi \etal \cite{choi2020part} propose a part-aware data augmentation that stochastically augments the partitions of objects by 5 basic augmentation methods. 
{LiDAR-Aug \cite{fang2021lidar} presents a rendering-based LiDAR augmentation framework to improve the robustness of 3D object detectors. LiDAR light scattering augmentation \cite{kilic2021lidar} and LiDAR fog stimulation \cite{hahner2021fog} utilize physics-based simulators to generate data corrupted by fog/snow/rain and then augment object detectors.}
Self-supervised pre-training \cite{yu2021point,zhang2021self} can also endow the model with resistance to augmentation-related transformations. Besides, denoising methods \cite{duan2021low,ning2018efficient,carrilho2018statistical} can remove the outliers in point clouds and thus potentially improve detectors' robustness. 
{Regarding module design, there are also some detectors specialized for resisting corruptions, \eg BtcDet \cite{xu2021behind} with the occupancy estimator for estimating occluded regions and Centerpoint \cite{yin2021center} with key-point detector for a flexible orientation regression.}
In this paper, we evaluate part-aware data augmentation and K-nearest-neighbors-based filtering methods for improving point cloud detectors against diverse common corruption patterns.

% \zhijie{Complete some todo blocks}

\section{Background}

\subsection{Point Cloud Detection}
% 公式化描述detection task
Point clouds detectors aim to detect objects of interest in point clouds in the format of \textit{bounding boxes} (BBoxes). Suppose a frame of point cloud data $\textbf{P}$ is a set of point $\textbf{p}=[x^p,y^p,z^p,r^p]$, where $(x^p,y^p,z^p)$ denotes its 3D location and $r^p$ denotes reflective intensity. Thus we can formulate the point cloud detection as: 
% 斜体，变量
% 整体， 文字和符号
% 向量，小写加粗
% 矩阵， 大写加粗
% c and s

\begin{equation}
    \begin{aligned}
        &\text{Det}(\textbf{P}) = \{\textbf{b}_i\}^{N} \\
        &\textbf{b}_i =[x_i, y_i, z_i, w_i, h_i, l_i, \theta_i, c_i, s_i]
    \end{aligned}
\end{equation}

% \begin{equation}
%     \mathrm{Det(\textbf{P}) = \{\textbf{b}_n\}^{N}}
% \end{equation}
% \begin{equation}
%     \mathrm{\textbf{b}_n=[x_n, y_n, z_n, w_n, h_n, l_n, \theta_n, c_n, s_n]}
% \end{equation}
where $\text{Det}(\cdot)$ represents the detector; $N$ is the number of detected BBoxes in $\textbf{P}$; $\textbf{b}_i$ denotes $i_{th}$ detected BBox in $\textbf{P}$, where $i=1, 2, \cdots ,N$; $(x_i, y_i, z_i)$ is the Cartesian coordinate of the center of $\textbf{b}_i$, $(w_i, h_i, l_i)$ is its dimensions, $\theta_i$ is its heading angle, $c_i$ is its classification label, and $s_i$ is its prediction confidence score. 

% In terms of \textbf{P}, there are different extracted representations used for the following feature abstraction. Besides, as for $\text{Det}(\cdot)$, there are different proposal architectures utilized to obtain object proposals.

%明显的分类引导（加粗）
% 四种 Representation intros and their pros and cons
{\noindent \bf Point cloud feature representation.} Representation for features used in point cloud detection includes 2D-view images, voxels, and raw points. By projecting point clouds into a 2D bird's eye view or front view, 2D-view-based 3D detectors can intuitively fit into a 2D image detection pipeline \cite{li2016vehicle,yang2018pixor}. However, 2D-view images could lose depth information \cite{qian20213d}, where the localization accuracy of the detector is affected. To efficiently acquire 3D spatial knowledge in large-scale point clouds, the ``voxelization'' operation is leveraged to partition unordered points into spatially and evenly distributed voxels \cite{yan2018second,ye2020hvnet}. After pooling interior features, those voxels are fed into a sparse 3D convolution backbone \cite{yan2018second} for feature abstraction. Given an appropriate voxelization scale, voxel-based representation is computationally efficient, but the quantization loss by voxelization is also inevitable \cite{qian20213d}. Different from the above methods, PointNet \cite{qi2017pointnet} and PointNet++ \cite{qi2017pointnet++} directly extract abstract features from raw points, which keeps the integrity of spatial context in point clouds. However, the point-based detectors are not cost-efficient for large-scale data \cite{qian20213d}. As a trade-off between the voxel-based and point-based methods, Point-voxel-based representations \cite{shi2020pv,he2020structure} possess the potential of fusing the high-efficient voxels and accurate-abstract points in feature abstraction. 
% 

% intro and pros and cons
{\noindent \bf Proposal architecture.} One-stage detectors \cite{yan2018second,zhang2021self} directly generate candidate BBoxes from the abstracted features. To improve candidate BBoxes' precision, two-stage detectors \cite{shi2020pv,xu2021behind} refine those BBoxes by region proposal network (RPN) and tailor them into unified size by region of interest (RoI) pooling before predicting output BBoxes. Compared with one-stage detectors, two-stage ones \cite{qian20213d} usually present more accurate localization but intuitively, are more computationally time-consuming.

% Even though strengths and weaknesses of different representations and proposal architectures have been explored \cite{arnold2019survey,qian20213d} in public datasets (\eg, KITTI \cite{geiger2013vision}, NuScenes \cite{caesar2020nuscenes}, and Waymo \cite{sun2020scalability}), their robustness performance under various common corruptions have not been comprehensively studied.

\subsection{Robustness Enhancement Solutions}

% 简要 intro DA and denoising effects
Several attempts have been made to enhance the robustness of point cloud detectors. In this paper, we select data augmentation and denoising methods to study their effects on improving point cloud detectors' robustness against common corruptions. 
%\zhijie{Rewrite the followings, use formal and scientific language with reference support} 
Data augmentation \cite{van2001art} is an effective way of increasing the amount of relevant data by slightly modifying existing data or newly creating synthetic data from existing data. Data augmentation on the point cloud \cite{choi2020part,chen2020pointmixup} provides detectors with a way to be trained with a larger dataset and thus potentially obtain more robust detectors. Different from data augmentation, denoising \cite{duan2021low,ning2018efficient} serves as a pre-process to detect and remove spatial outliers in point clouds, which can reduce the effects of noisy point cloud data. 

% \section{Physical-aware Common Corruptions for Point Cloud Detector Benchmark}
\section{Physical-aware Robustness Benchmark for Point Cloud Detection}

% \subsection{Overview}
% % wenti and the goal of benchmark.
% %       两个问题：1）
% % 4.1， 4.2， 4.3
% % simulation kit
% % dataset
% % evaluation metric
% % detectors and enhancing 
% A robustness benchmark is indispensable to explore the affects of different corruptions on point cloud detection. Given collecting LiDAR point clouds
% affected by common corruptions in the real world  is of vast consumption of time, labor, and material resources, we utilize corruption simulation to build the corrupted dataset on the basis of existing clean data. To ensure the fidelity of corruption simulations, we shape a kit of the physical-aware simulation methods to generate degraded point clouds under different common corruptions. Section 4.2 shows the details on the physical-aware simulation kit and the strategy of constructing dataset. In Section 4.3, we design two metric sets of overall accuracy and detection bug for comparison experiments. In the end, we introduce the targets of benchmarks in Section 4.4. 

%简化
We propose the first robustness benchmark of point cloud detectors against common corruption patterns. We first introduce different corruption patterns collected for this benchmark and dataset in Section~\ref{sec:benchmark}. Then we propose the evaluation metrics used in our benchmark in Section~\ref{sec:metrics}. Finally, we introduce the subject-object detection methods and robustness enhancement methods selected for this benchmark in Section~\ref{sec:detectors}.
% To explore the effects of different corruptions on point cloud detection, we propose a robustness benchmark constructed by physical-aware corruption simulation methods: first, we shape a kit of the physical-aware simulation methods and generate the benchmark dataset containing degraded point clouds under different corruptions in Section 4.1; then we design two metric sets of overall accuracy and detection bug for comparison experiments in Section 4.3; finally, we introduce the targets of robustness benchmark in Section 4.3.

\subsection{Physical-aware Corrupted Dataset Construction}
\label{sec:benchmark}
% corruptions 分类
% 强调physics by 指出原因

% 加一段说明，现有sensor调研，当然这是initial 努力

After the literature investigation in Section \ref{sec: lidar_perception}, we summarize 25 corruption patterns in Table \ref{tab: corruption details} and categorize them into 4 categories based on presentations of common corruptions: \textit{weather, noise, density, and transformation}. On the other hand, we also divide common corruption patterns into the \textit{scene-level} and the \textit{object-level}. 
{As an initial effort, the dataset covers representative but not all corruptions, and we encourage continuous work with more diverse corruptions considered in the future. 

The simulation of corruptions implemented in the paper mainly operates on the spatial locations and the reflection intensity of points in the point cloud. Those point-targeting operations are equivalent to the perturbations of the real-world corruptions on the LiDAR point cloud and have been widely utilized in the simulation-related studies, as in noise-related \cite{ma2012analysis,Texas2016lidar,ren2022benchmarking,sun2022benchmarking,choi2020part}, density-related \cite{ren2022benchmarking,sun2022benchmarking,choi2020part,cheng2020improving,xu2021behind}, and transformation-related \cite{ren2022benchmarking,sun2022benchmarking,wang2020train,morin2021simulated,cheng2020improving}.}
Next, We briefly introduce each corruption pattern in the following (refer to Appendix \ref{sec: corruption_simulation_implementation} for detailed implementations and visualizations).

% \paragraph{Weather Corruptions}
% cause and effect
% the intro to simulator and fidelity validation.
{\noindent \textbf{Weather corruptions:}} LiDAR is sensitive to adversarial weather conditions, such as rainy, snowy, and foggy \cite{rasshofer2011influences}. Dense droplets of liquid or solid water dim the reflection intensity and reduce the signal-to-noise ratio (SNR) of received lights. Floating droplets can also reflect and fool sensors with false alarms. Both effects, in some cases, can significantly affect  the detectors. To simulate three weather corruptions: \{\textit{rain, snow, fog}\}, we adopt LiDAR light scattering augmentation (LISA) \cite{kilic2021lidar} as a simulator for rain and snow and LiDAR fog stimulation (LFS) \cite{hahner2021fog} as a fog simulator. 

% fidelity testing: mechanism and results.
To verify the naturalness of weather simulation, we train weather-oriented PointNet-based classifiers with datasets collected in real snowy and foggy weather. Then, we leverage the classification accuracy of those trained classifiers testing on simulated data to measure the similarity of simulated data to real data. As shown in Table \ref{tab: weather classification}, the testing accuracy 97.13\% and 92.60\% of trained weather classifiers on simulated snow data and fog data show that the simulated snow data and fog data are highly similar to the real data (refer to Appendix \ref{sec: naturalness validation for weather} for detailed experiment settings). 

\begin{table}[t]
\centering
\caption{{Classification on real and simulated data}}
\label{tab: weather classification}
\resizebox{0.48\textwidth}{!}{
\begin{tabular}{c|ccc|ccc}
\toprule
\multirow{2}{*}{\textbf{Corruption}} & \multicolumn{3}{c|}{\textbf{Training}} & \multicolumn{3}{c}{\textbf{Testing}} \\ \cmidrule(r){2-7}
              & dataset & size & \textit{val} accuracy & dataset & size & \textit{test} accuracy \\ \midrule
\textit{snow} & Boreas & 24292 & 99.11$\%$   & KITTI & 14962 & 97.13$\%$ \\
\textit{fog}  & STF    & 1787  &  80.00$\%$  & KITTI & 14962 & 92.60$\%$ \\ 
\bottomrule
\end{tabular}
}
\end{table}

\begin{figure}[b]
  \centering
  \vspace{-10pt}
  \includegraphics[width=1\linewidth]{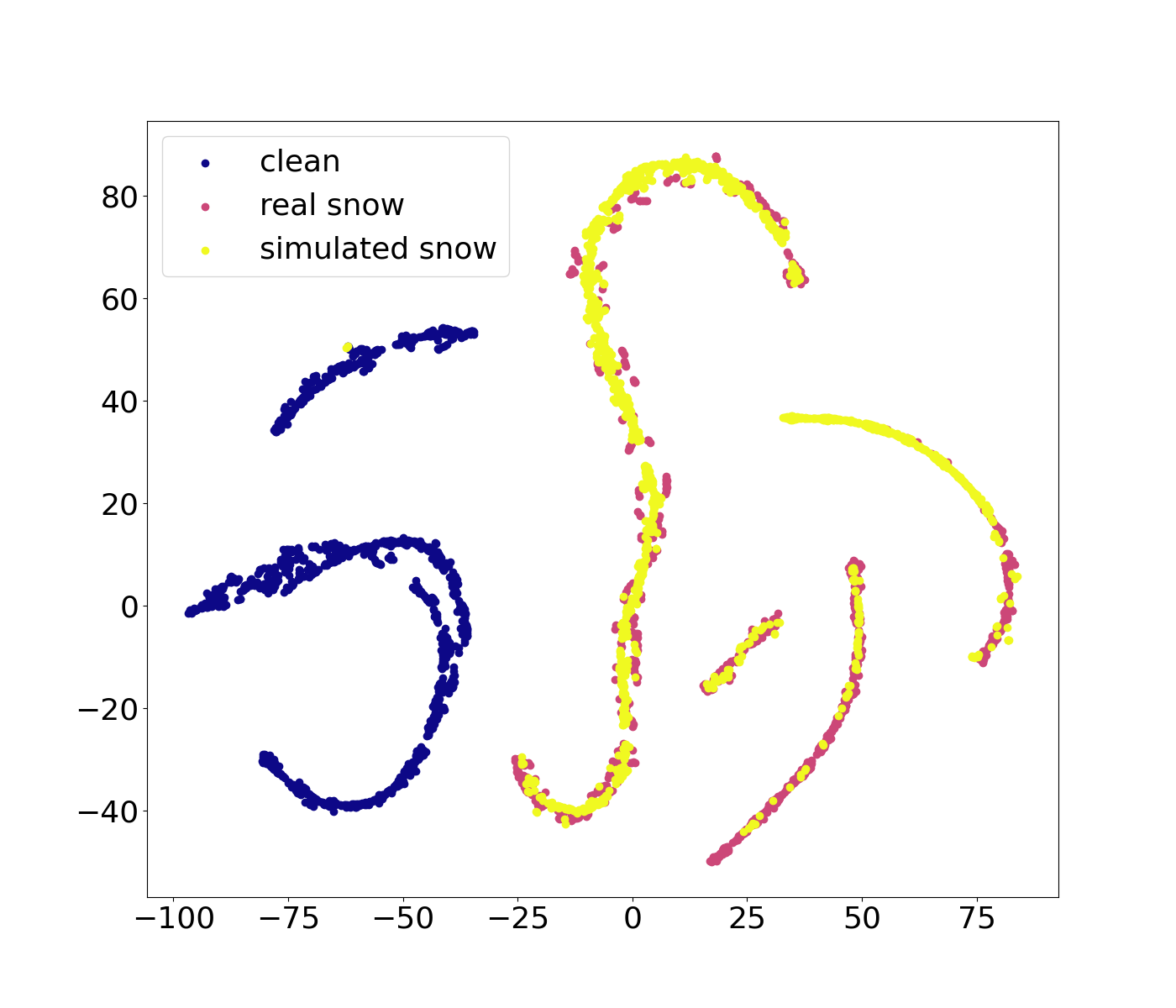}
  \caption{{Feature visualization of the snow classification by T-SNE}}
  \label{fig: snow_classification_TSNE}
\end{figure}

\begin{figure}[t]
  \centering
  \vspace{-30pt}
  \includegraphics[width=1\linewidth]{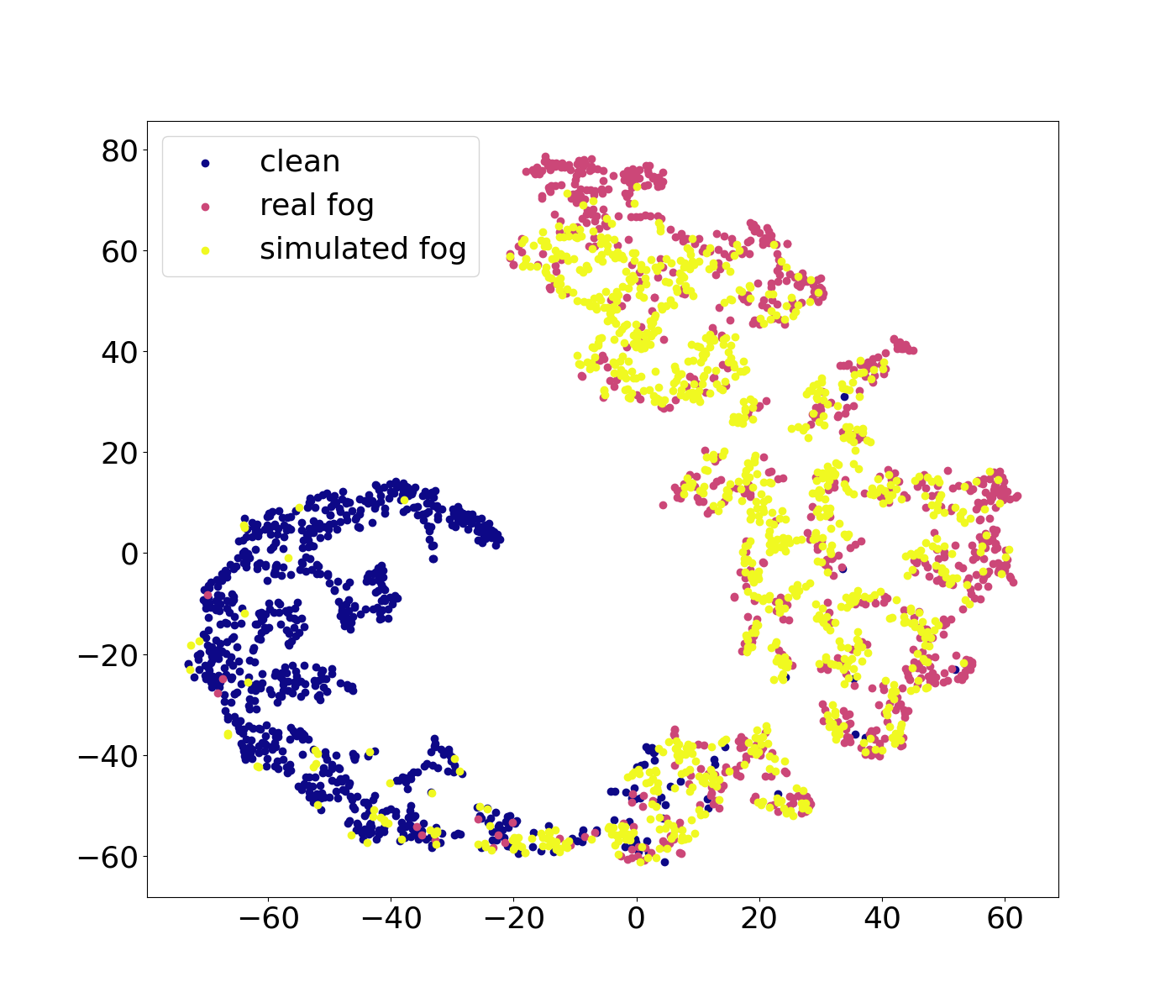}
  \caption{{Feature visualization of the fog classification by T-SNE}}
  \label{fig: fog_classification_TSNE}
\end{figure}

{We further analyze the similarity of distribution of real and simulated corrupted data. Specifically, we extract the high-level features from the trained classifier. Then, we utilize T-SNE \cite{van2008visualizing} to reduce the dimensionality of acquired features to 2 and visualize these 2D features. As shown in Figure \ref{fig: snow_classification_TSNE} and \ref{fig: fog_classification_TSNE}, the distributions of the real and simulated corruptions are significantly similar. We further quantitatively measure the distance between the feature distribution of clean data, simulated \textit{snow/fog}, and real \textit{snow/fog}, as shown in Table \ref{tab: MMD}. 
The maximum mean discrepancy (MMD) \cite{gretton2006kernel} results reveal that the simulated \textit{snow/fog} is close to the real \textit{snow/fog}, respectively, while not close to the clean data.  
}

\begin{table}[t]
\centering
\caption{{MMD distances among clean, real corrupted, and simulated corrupted 2D features transformed by T-SNE}}
\label{tab: MMD}
\resizebox{0.45\textwidth}{!}{%
\begin{tabular}{@{}c|c|c|c@{}}
\toprule
 & \textbf{Real \textit{vs} Simulated } & \textbf{Simulated  \textit{vs} Clean} & \textbf{Real \textit{vs} Clean} \\ \midrule
\textbf{snow} & {\color{red} 0.0549} & 0.1446 & 0.1445 \\
\textbf{fog} & {\color{red} 0.0302} & 0.1212 & 0.1295 \\ \bottomrule
\end{tabular}%
}
\end{table}

\begin{figure*}[h]
  \centering
  \includegraphics[width=0.8\linewidth]{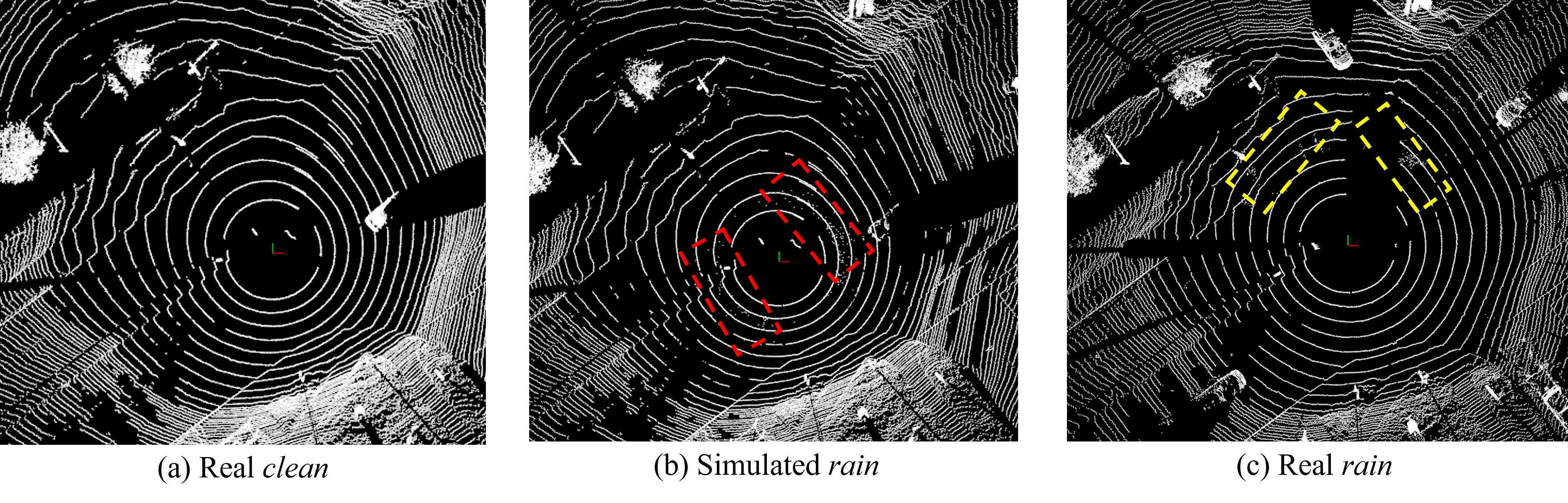}
  \caption{Comparison between real \textit{rain} and simulated \textit{rain} (red and yellow boxes contain the false points in the simulated and real \textit{rain}, respectively; the data of real \textit{clean} and real \textit{rain} from Boreas were sampled at the same location; the simulated \textit{rain} data was augmented on the basis of the real \textit{clean} data)}
  \label{fig: rain_comparison}
\end{figure*}

Regarding \textit{rain} corruption, we find the effects of rain droplets on point clouds are too subtle to be caught by classifiers, as shown in Figure \ref{fig: rain_comparison}. Alternatively, we visualize simulated and real point clouds and qualitatively verify the high similarity between simulated and real point clouds (see more comparisons in Appendix \ref{sec: naturalness validation for weather}). 

% \paragraph{Noise Corruptions}
% for ranging in {scene-level, object-level}
%    cause and effect (note the difference of the scene-level and object-level)
%    the implements of simulator.
{\noindent \textbf{Noise corruptions:}} Noise commonly exists in point cloud signals \cite{carrilho2018statistical, duan2021low}. Scene-level factors (\eg, strong illumination \cite{villa2021spads}, limited ranging accuracy of sensors, and sensor vibration \cite{ma2012analysis,wang2021simultaneous}) could increase the variance of ranging or extend the positioning bias. Floating particles, \eg, dust \cite{mona2012lidar}, could cause the background noise in point clouds. Hence, we collect 5 scene-level noise corruptions: 
\{\textit{uniform$\_$rad, gaussian$\_$rad, impulse$\_$rad}\} add uniform, Gaussian, impulse noise on the spherical coordinates of points in point clouds; \{\textit{upsample}\} randomly upsamples points nearby original points in point clouds; \{\textit{background}\} uniformly randomly samples points within the spatial range of point clouds. Besides scene-level effects, object-related factors could cause noise in LiDAR points, \eg, dark color \cite{bolkas2018effect} and coarse surface. Thus, we formulate 4 object-level corruptions: \{\textit{uniform, gaussian, impulse}\} add uniform, Gaussian, impulse noise on the Cartesian coordinates of points of objects; \{\textit{upsample}\} upsamples points nearby original points of objects.

% \paragraph{Density Corruptions}
% similar to last one
{\noindent \textbf{Density corruptions:}} {The density-related corruptions refer to the corruption patterns that change the global or local density distribution of LiDAR point clouds.}
For instance, the global static density of points in LiDAR varies due to different amounts of scanning layer (\eg, 32 or 64). Besides, inter-object occlusion and random signal loss \cite{xu2021behind} could remove points randomly. We hence propose 5 corruptions: \{\textit{cutout}\} cuts out the sets of locally gathering points; \{\textit{local$\_$dec, local$\_$inc}\} locally decrease or increase the density of points; \textit{beam$\_$del, layer$\_$del} randomly delete points or layers of points in point clouds. In terms of object-level factors, dark-color cover \cite{bolkas2018effect} and transparent materials (\eg, glasses and plastics) of objects can affect the point density of objects. Hence, at the object level, we also propose a set of corruptions: \{\textit{cutout, local$\_$dec, local$\_$inc}\}, affecting the point density of objects.

% cause and effect (note the difference of the scene-level and object-level)
% the implements of simulator.

% \paragraph{Transformation Corruptions}
% cause and effect
% the implements of simulator.
{\noindent \textbf{Transformation corruptions:}} In the scenario of autonomous driving, shapes of objects within one class could be various (\eg, flat sports cars and round vintage cars \cite{wang2020train}, bending and walking pedestrians \cite{wong2020efficient}). Those long-tail data could potentially be recognized wrong. Besides, dynamic changes in heading directions and locations of objects \cite{morin2021simulated} could potentially affect the positioning accuracy of detectors. Hence, we formulate 5 corruptions: \{\textit{translation, rotation}\} change locations and heading directions of objects to a milder degree, \ie, $<1m$ and $<10^\circ$; \{\textit{shear}\} \cite{chen20003d} and \{\textit{scale}\}, as linear deformations, slant and scale points of objects; \{\textit{FFD}\} adopts free-form deformation (FFD) \cite{sederberg1986free} to distort the point shape of an object in a nonlinear manner. 

% intro to kitti and reason that we choose it
{\noindent \bf Dataset selection.} As one of the most popular benchmarks in autonomous driving, KITTI \cite{geiger2013vision} contains 7481 training samples covering 8 object classes. {Unlike other datasets in Table \ref{tab: summary_dataset}, the data in KITTI are mostly collected under clean conditions and also have a relatively simple annotation format, which makes it a good option for conducting comparative experiments. We also encourage the future extension to other real or synthesized datasets.}
% strategy of construction
%   6 severity levels
%   corrupt val set and leave train behind
To simulate various levels of severity in the real world, we set 6 severity levels for each corruption (considering ``clean'' as level 0).

\subsection{Evaluation Metrics}
\label{sec:metrics}

To quantify the robustness performance of detectors, we design the following evaluation metrics from two perspectives: (1) detection accuracy and (2) number of bugs triggered.

{\noindent \bf Overall accuracy.}
% intro to OA
For each test, we use the overall accuracy (OA), by taking the average of APs (\textit{average precision}) at three difficulty levels (\ie, ``Easy'', ``Moderate'', and ``Hard''). And we follow the common settings of IoU thresholds \{Car: 0.7, Pedestrian: 0.5, Cyclist: 0.5\} to search for the true positive detections in AP and recall calculation. 

% intro to CE and mCE
For every corruption, we calculate corruption error (CE) to measure performance degradation according to OA by:% $CE_{c,s}^m = OA_{clean}^m - OA_{c, s}^m$,
\begin{equation}
    \mathrm{CE_{c,s}^m = \text{OA}_\text{clean}^m - \text{OA}_{c, s}^m}
\end{equation}
where $OA_{c,s}^m$ is the overall accuracy of detector $m$ under corruption $c$ of severity level $s$ (exclude ``clean'', \ie, severity level 0) and $clean$ represent the clean data. For detection $m$, we can calculate the mean CE (mCE) for each detector by: % $mCE^m = \frac{\sum_{s=1}^{5} \sum_{c=1}^{25} CE_{c,s}^m}{5\cdot C}$.
\begin{equation}
    \mathrm{mCE^m = \frac{\sum_{s=1}^{5} \sum_{c=1}^{25} CE_{c,s}^m}{5C}}
\end{equation}

\begin{table*}[ht]
\centering
\caption{{AP(\%) of all detectors under clean observations (at the severity level of 0)}}
\label{tab: AP_detector_clean}
% \resizebox{\linewidth}{!}{%
\begin{tabular}{@{}ccccccccc@{}}
\toprule
 & \textbf{PVRCNN} & \textbf{PointRCNN} & \textbf{SECOND} & \textbf{BtcDet} & \textbf{VoTr-SSD} & \textbf{VoTr-TSD} & \textbf{SESSD} & \textbf{Centerpoint} \\ \midrule
\textbf{Car}        &  86.77 & 82.82 & 83.67 & 87.32 & 81.04 & 86.39 & 86.44 & 82.14 \\
\textbf{Pedestrian} & 60.61 & 52.34 & 52.15 & - & - & - & - & 49.32 \\
\textbf{Cyclist}    & 76.42 & 77.60 & 68.51 & - & - & - & - & 68.58 \\ \bottomrule
\end{tabular}%
% }
\end{table*}

\begin{table*}[ht]
\centering
\caption{ $CE_{AP}(\%)$ of different detectors under different corruptions on \textit{Car} detection (the \colorbox{green!40}{\color{black}green} cell stands for the lowest $CE_{AP}$ among detectors given a certain corruption and the {\colorbox{yellow!60}{yellow}} cell for the average $mCE_{AP}$)}
\label{tab: CEs of different detectors and CCs}
\resizebox{\textwidth}{!}{%
\scriptsize
\begin{tabular}{c|c|c|cccccccc|c}
\toprule
\multicolumn{3}{c|}{\multirow{2}{*}{\textbf{Corruption}}} & \multicolumn{1}{c|}{\textbf{Point-voxel}} & \textbf{Point} & \multicolumn{6}{|c|}{\textbf{Voxel}} & \multicolumn{1}{c}{\multirow{2}{*}{\textbf{Average}}} \\ \cmidrule(r){4-11} 
\multicolumn{3}{c|}{} & \multicolumn{1}{c|}{PVRCNN} & \multicolumn{1}{c|}{PointRCNN} & SECOND & BtcDet & VoTr-SSD & VoTr-TSD & SE-SSD & Centerpoint & \\  \midrule
\multirow{13}{*}{\textbf{Scene-level}} & \multirow{3}{*}{\textbf{Weather}} & \textit{rain} & 25.11 & 23.31 & \cellcolor{green!40} 21.81 & 31.07 & 28.17 & 26.77 & 29.51 &  25.83 & 26.45 \\
 & & \textit{snow} & 44.23 & 37.74 & \cellcolor{green!40} 34.84 & 54.07 & 54.10 & 52.18 & 49.19 & 38.74 & 45.64 \\
 & & \textit{fog}  & 1.59  & 3.52  & 1.60  & 1.81  & 1.77  & 2.02  &  1.59  & \cellcolor{green!40} 1.11 & 1.88  \\ \cmidrule(r){2-12} 
 & \multirow{5}{*}{\textbf{Noise}} & \textit{uniform\_rad}  & 10.19 & 8.32  & 9.51  & 9.13  & \cellcolor{green!40} 3.79  & 4.11  & 9.34 & 8.15 & 7.82  \\ 
 &  & \textit{gaussian\_rad} & 13.02 & 9.98  & 12.13 & 10.83 & \cellcolor{green!40} 4.84  & 5.18  & 11.02 &  10.17 & 9.65  \\
 &  & \textit{impulse\_rad}  & 2.20  & 3.86  & 2.23  & 2.50  & 2.25  & 3.57  & \cellcolor{green!40} 1.18  &  1.86 & 2.46  \\
 &  & \textit{background}    & 2.93  & 6.49  & 2.41  & \cellcolor{green!40} 1.82  & 4.59  & 3.68  & 2.14  &  1.86 & 2.46  \\
 &  & \textit{upsample}      & 0.81  & 1.84  & \cellcolor{green!40} 0.31  & 0.95  & 0.37  & 0.71  & 0.55  &  0.46 & 0.75  \\ \cmidrule(r){2-12} 
 & \multirow{5}{*}{\textbf{Density}} & \textit{cutout} & 3.75  & 3.97  & 4.27  & 3.99  & 4.51  & \cellcolor{green!40} 3.59  & 4.26  &  4.11 & 4.06  \\ 
 &  & \textit{local\_dec}    & 14.04 & -     & 13.88 & 14.55 & 14.44 & \cellcolor{green!40} 12.50 & 17.04 &  14.64 & 14.44 \\
 &  & \textit{local\_inc}    & 1.40  & 3.34  & 1.33  & 2.20  & 1.66  & 1.69  & \cellcolor{green!40} 0.90  &  0.95 & 1.68  \\
 &  & \textit{beam\_del}     & 0.58  & 0.79  & 0.73  & 0.88  & 0.80  & 0.53  & 1.07  & \cellcolor{green!40} 0.47 & 0.73  \\
 &  & \textit{layer\_del}    &  2.94  & 3.46  & 3.10  & 3.39  & 3.29  & 3.16  & 3.37  &  \cellcolor{green!40}2.67 & 3.17  \\ \midrule
\multirow{12}{*}{\textbf{Object-level}} & \multirow{4}{*}{\textbf{Noise}} & \textit{uniform} & 15.44 & 12.95 & 9.48 & 12.60 & \cellcolor{green!40} 2.76 & 4.81 & 6.99 & 6.51 & 8.94 \\ 
 &  & \textit{gaussian}      & 20.48 & 17.62 & 12.98 & 17.05 & \cellcolor{green!40} 4.72  & 7.46  & 9.56  &  9.49 & 12.42 \\
 &  & \textit{impulse}       & 3.30  & 4.70  & 2.53  & 4.07  & 2.88  & 4.29  &  2.20  &  \cellcolor{green!40} 2.11 & 3.26  \\
 &  & \textit{upsample}      & 1.12  & 1.95  & 0.67  & 1.33  & \cellcolor{green!40} 0.08  & 0.40  & 0.22  &  0.16 & 0.74  \\ \cmidrule(r){2-12} 
 & \multirow{3}{*}{\textbf{Density}} & \textit{cutout} & 15.81 & 15.62 & 14.99 & 15.62 & 15.07 & 16.09 & 16.51 & \cellcolor{green!40} 14.06 & 15.47 \\ 
 &  & \textit{local\_dec}    & 14.38 & 14.16 & 13.23 & 14.26 & 12.66 & 14.41 & 15.08 & \cellcolor{green!40} 12.52 & 13.84 \\
 &  & \textit{local\_inc}    & 13.93 & 14.19 & 13.74 & 13.56 & 11.34 & 13.05 & \cellcolor{green!40} 11.03 &  11.64 & 12.81 \\ \cmidrule(r){2-12} 
 & \multirow{5}{*}{\textbf{Transformation}} & \textit{shear} & \cellcolor{green!40} 37.27 & 40.96 & 40.35 & 41.37 & 39.52 & 37.85 & 40.35 & 40.00 & 39.71 \\ 
 &  & \textit{FFD}           & \cellcolor{green!40} 32.42 & 38.88 & 33.15 & 36.77 & 33.14 & 34.26 & 37.96 &  32.86 & 34.93 \\
 &  & \textit{rotation}      & 0.60  & 0.47  & 0.31  & 0.97  & 0.39  & 0.75  & \cellcolor{green!40} 0.27  &  0.38 & 0.52  \\
 &  & \textit{scale}       & \cellcolor{green!40} 5.78  & 8.13  & 6.96  & 5.81  & 8.53  & 6.50  & 6.53  &  7.50 & 6.97  \\
 &  & \textit{translation}   & 3.82  & 3.03  & 3.24  & 4.58  & 4.88  & 5.34  & \cellcolor{green!40} 1.37  &  3.91 & 3.77  \\ \midrule
\multicolumn{3}{c|}{mCE}   & 11.49 & 11.64 & 10.39 & 12.21 & 10.42 & 10.60 & 11.17 & \cellcolor{green!40} 10.09 & \cellcolor{yellow!60} 11.01 \\ \bottomrule
\end{tabular}
}
\end{table*}

{\noindent \bf Detection bug.}
% why use bugs
% what kind of bugs
% intro BR and CR
There are various bugs existing in the pipeline of point cloud detection, such as annotation errors, run-time errors, detection bugs. In this paper, we focus on the bugs in detection results. Specifically, we're interested in false detection, false classification, and missed detection:
\begin{itemize}[leftmargin=*]
    \item False detection (FD) on detection BBoxes: maximum IoU > 0 with correct classification w.r.t. ground-truth BBoxes;
    \item False classification (FC) on detection BBoxes: maximum IoU > 0 with false classification w.r.t. ground-truth BBoxes;
    \item Missed detection (MD) on detection BBoxes: maximum IoU = 0 w.r.t. ground-truth BBoxes. 
\end{itemize}

Correspondingly, the bug rates (BRs) are calculated by:% $BR_{*} = {N_{*}}/{N_{det}}$,
\begin{equation}
    \mathrm{BR_{*} = \frac{N_{*}}{N_{det}}} 
\end{equation}
where ${*}$ stands for FD, FC, and MD; $N_{*}$ is the number of objects of ${*}$; $N_{det}$ is the number of detected objects. 

To measure the increase of BR after being affected by common corruptions, we calculate corruption risk (CR) and the mean CR (mCR) for detector $m$ by 

% by $CR_{*, c,s}^m = BR_{*, c, s}^m-BR_{*, clean}^m$ and the mean CR (mCR) for detector $m$ by $mCR^m_{*} = \frac{\sum_{s=1}^{5} \sum_{c=1}^{C}  CR_{*, c,s}^m}{5\cdot C}$, 
\begin{equation}
    \mathrm{CR_{*, c,s}^m = BR_{*, c, s}^m-BR_{*, clean}^m}
\end{equation}
\begin{equation}
    \mathrm{mCR^m_{*} = \frac{\sum_{s=1}^{5} \sum_{c=1}^{C}  CR_{*, c,s}^m}{5C}}
\end{equation}
where $BR_{*, c, s}^m$ is the $BR_{*}$ of detector $m$ under corruption $c$ of severity level $s$.

\subsection{Benchmark Subjects}
\label{sec:detectors}

{\noindent \bf Point cloud detectors.}
% what detectors choose, and intros
% why we choose them, 
For benchmarking point cloud detection, we select 8 representative detectors: SECOND \cite{yan2018second}, PointRCNN \cite{shi2019pointrcnn}, PVRCNN \cite{shi2020pv}, BtcDet \cite{xu2021behind}, VoTr-SSD, VoTr-TSD \cite{mao2021voxel}, Centerpoint \cite{yin2021center}, and SE-SSD \cite{zheng2021se} to cover different kinds of feature representations and proposal architectures. We show the detailed taxonomy in Table~\ref{tab:Types of detectors}.

{\noindent \bf Data augmentation and denoising methods.}
% intro to PA-DA and KNN-OR
% Several attempts have been made to enhance the robustness of point cloud detectors, \eg, data augmentation and denoising. 
In this paper, we study the effectiveness of data augmentation and denoising methods for improving detectors' robustness against corruption. For data augmentation, we choose the part-aware data augmentation (PA-DA) method \cite{choi2020part}. For denoising, we adopt K-nearest-neighbors-based outlier removing (KNN-OR) \cite{carrilho2018statistical} to remove the outliers out with 3 times the standard deviation of distance distribution within the cluster of 50 points.

\begin{table}[t]
\centering
\caption{Subject point cloud detectors.}
\label{tab:Types of detectors}
\resizebox{0.95\linewidth}{!}{%
\scriptsize
\begin{tabular}{l|ccc|cc}
\toprule
\multicolumn{1}{c}{\multirow{2}{*}{\textbf{Detectors}}} & \multicolumn{3}{|c}{\textbf{Representations}} & \multicolumn{2}{|c}{\textbf{Proposal Architectures}} \\ \cmidrule(r){2-6} 
\multicolumn{1}{c|}{} & point & voxel & point-voxel & one-stage & two-stage \\ \midrule
\textbf{SECOND} &  & \checkmark &  & \checkmark &  \\
\textbf{PointRCNN} & \checkmark &  &  &  & \checkmark \\
\textbf{PVRCNN} &  &  & \checkmark &  & \checkmark \\
\textbf{BtcDet} &  & \checkmark &  &  & \checkmark \\
\textbf{VoTr-SSD} &  & \checkmark &  & \checkmark &  \\
\textbf{VoTr-TSD} &  & \checkmark &  &  & \checkmark \\
\textbf{SE-SSD} &  & \checkmark &  & \checkmark & \\ 
\textbf{Centerpoint} &  & \checkmark &  &  & \checkmark \\ 
\bottomrule
\end{tabular}%
}
\end{table}

\section{Experiments and Analysis}

% With the help of the proposed benchmark, we conduct empirical study to answer following questions:
% \begin{itemize}[leftmargin=*]
%     \item How do the common corruption patterns affect the point cloud detector’s performance?
%     \item How does the design of a detector affect its robustness under common corruptions?
%     \item What kind of bugs exist in point cloud detectors under common corruptions?
%     \item How do the existing robustness enhancement methods improve point cloud detectors?
% \end{itemize}
% These questions will be answered in Sections \ref{sec: main_Effects of Common Corruptions to Point Cloud Detectors}, \ref{sec: main_Reacts of Detector Designing to Common Corruptions}, \ref{sec: main_Detection Bugs in Detectors under Common Corruptions}, and \ref{sec: main_Robustness Enhancement by Data Augmentation and Denoising}. 

\subsection{Experimental Set-ups}
\label{sec:experiment_setting}
% experiment settings
For a fair comparison, each detector in Table~\ref{tab:Types of detectors} is trained with the clean training set of KITTI, following the training strategy in each paper, and evaluated with corrupted validation sets of KITTI. 
All detectors are executed based on the open-source codes released on GitHub, as shown in Table \ref{tab: det_open} in the Appendix~\ref{sec: appendix_empirical_study}. The configuration files and pre-trained checkpoints can be found in Table \ref{tab: det_open}.

The training and evaluation are all executed on the NVIDIA RTX A6000 GPU with a memory of 48GB. The batch size of each detector is optimized to reach the limit of GPU memory. In the experiments of robustness enhancement, data augmentation is adopted to augment the clean \textit{train} dataset before training and the denoising directly processes the \textit{val} data during the testing stage. Note that, since only detection of "Car" is available for all detectors, as shown in Table \ref{tab: AP_detector_clean}, the following evaluation will mainly focus on detected results in the "Car" category. We encourage readers refer to the Appendix \ref{sec: appendix_empirical_study} for complete evaluation results, \eg, about "Pedestrian".

\begin{table*}[ht]
\centering
\caption{$CE_{recall}(\%)$ of different detectors under different corruptions on \textit{Car} detection (the {\colorbox{green!40} {green}} cell with $CE$ of over $15\%$)}
\label{tab: CE_Recalls of different detectors and CCs}
\resizebox{\textwidth}{!}{%
\begin{tabular}{c|c|c|cccccccc|c}
\toprule
\multicolumn{3}{c|}{\textbf{Corruption}} & \textbf{PVRCNN} & \textbf{PointRCNN} & \textbf{SECOND} & \textbf{BtcDet} & \textbf{VoTr-SSD} & \textbf{VoTr-TSD} & \textbf{SE-SSD} & \textbf{Centerpoint} & \textbf{Average} \\ \midrule
\multirow{13}{*}{\textbf{Scene-level}} & \multirow{3}{*}{\textbf{Weather}} & \textit{rain} & 24.50 & 23.67 & 20.92 & 29.69 & 23.46 & 24.98 & 27.10 & 24.84 & \cellcolor{green!40} 24.90 \\
 &  & \textit{snow} & 36.23 & 32.72 & 29.19 & 43.32 & 37.11 & 40.26 & 38.32 & 32.80 & \cellcolor{green!40} 36.24 \\
 &  & \textit{fog} & 4.22 & 5.30 & 3.14 & 2.36 & 2.45 & 2.43 & 2.28 & 3.47 & 3.21 \\ \cmidrule(r){2-12} 
 & \multirow{5}{*}{\textbf{Noise}} & \textit{uniform\_rad} & 14.23 & 12.91 & 13.67 & 9.20 & 3.70 & 5.12 & 9.33 & 11.31 & 9.93 \\
 &  & \textit{gaussian\_rad} & 17.74 & 15.67 & 17.46 & 11.30 & 4.81 & 6.39 & 11.29 & 14.35 & 12.38 \\
 &  & \textit{impulse\_rad} & 4.17 & 4.47 & 4.32 & 3.93 & 3.86 & 5.02 & 2.11 & 4.05 & 3.99 \\
 &  & \textit{background} & 3.26 & 8.66 & 2.84 & 2.21 & 4.64 & 4.73 & 2.59 & 2.42 & 3.92 \\
 &  & \textit{upsample} & 0.94 & 2.58 & 1.20 & 0.93 & 0.94 & 0.68 & 0.80 & 0.76 & 1.10 \\ \cmidrule(r){2-12} 
  & \multirow{5}{*}{\textbf{Density}} & \textit{cutout} & 4.87 & 4.61 & 5.55 & 4.05 & 4.37 & 3.55 & 4.45 & 5.55 & 4.62 \\
 &  & \textit{local\_dec} & 15.56 & - & 14.76 & 13.04 & 11.78 & 11.10 & 14.06 & 16.19 & 13.78 \\
 &  & \textit{local\_inc} & 1.58 & 2.62 & 1.59 & 1.66 & 1.70 & 1.48 & 1.21 & 1.56 & 1.68 \\
 &  & \textit{beam\_del} & 0.92 & 0.77 & 1.13 & 0.95 & 0.94 & 0.63 & 1.34 & 1.11 & 0.97 \\
 &  & \textit{layer\_del} & 3.48 & 3.37 & 3.59 & 3.12 & 3.13 & 2.74 & 3.37 & 3.78 & 3.32 \\ \midrule
\multirow{12}{*}{\textbf{Object-level}} & \multirow{4}{*}{\textbf{Noise}} & \textit{uniform} & 9.60 & 10.19 & 6.74 & 9.22 & 2.65 & 3.67 & 7.05 & 5.82 & 6.87 \\
 &  & \textit{gaussian} & 12.69 & 13.02 & 9.16 & 12.05 & 3.84 & 5.18 & 9.06 & 7.78 & 9.10 \\
 &  & \textit{impulse} & 2.11 & 3.14 & 1.88 & 2.03 & 1.99 & 1.94 & 1.95 & 1.86 & 2.11 \\
 &  & \textit{upsample} & 0.77 & 1.71 & 0.96 & 1.00 & 0.44 & 0.26 & 0.44 & 0.57 & 0.77 \\ \cmidrule(r){2-12} 
 & \multirow{3}{*}{\textbf{Density}} & \textit{cutout} & 22.21 & 20.90 & 21.61 & 17.61 & 16.27 & 17.44 & 17.67 & 21.21 & \cellcolor{green!40} 19.36 \\
 &  & \textit{local\_dec} & 19.99 & 18.73 & 19.04 & 15.97 & 14.11 & 15.52 & 16.26 & 18.96 & \cellcolor{green!40} 17.32 \\
 &  & \textit{local\_inc} & 10.58 & 10.95 & 10.78 & 9.30 & 7.63 & 8.11 & 7.89 & 9.33 & 9.32 \\ \cmidrule(r){2-12} 
  & \multirow{5}{*}{\textbf{Transformation}} & \textit{shear} & 22.13 & 25.19 & 25.06 & 23.26 & 22.14 & 20.61 & 21.70 & 23.95 & \cellcolor{green!40} 23.00 \\
 &  & \textit{FFD} & 17.53 & 21.51 & 18.41 & 19.43 & 16.74 & 17.26 & 18.70 & 18.43 & \cellcolor{green!40} 18.50 \\
 &  & \textit{rotation} & 0.49 & 0.42 & 0.4 & 0.63 & 0.4 & 0.42 & 0.53 & 0.26 & 0.44 \\
 &  & \textit{scaling} & 5.1 & 5.95 & 5.95 & 4.56 & 6.09 & 4.77 & 4.82 & 5.86 & 5.39 \\
 &  & \textit{translation} & 3.79 & 3.42 & 3.78 & 4.69 & 5.32 & 4.66 & 2.39 & 4.26 & 4.04 \\ \midrule
\multicolumn{3}{c|}{mCE} & 10.35 & 10.52 & 9.73 & 9.82 & 8.02 & 8.36 & 9.07 & 9.62 & 9.45 \\
\bottomrule
\end{tabular}
}
\end{table*}

\begin{table}[h]
\centering
\caption{$CE_{AP}(\%)$ under different severity levels of different common corruptions on \textit{Car} detection ({\colorbox{yellow!60}{yellow}} cells for the $CE_{AP}$ under \textit{rain})}
\label{tab: CEs of different CCs and severities}
\resizebox{0.48\textwidth}{!}{%
\begin{tabular}{c|c|c|ccccc|c}
\toprule
\multicolumn{3}{c|}{\textbf{Corruption}} & \textbf{1} & \textbf{2} & \textbf{3} & \textbf{4} & \textbf{5} & \textbf{Average} \\ \midrule
\multirow{13}{*}{\rotatebox{90}{\textbf{Scene-level}}} & \multirow{3}{*}{\textbf{Weather}} & \textit{rain} & \cellcolor{yellow!60} 27.11 & \cellcolor{yellow!60}26.80 & \cellcolor{yellow!60}25.44 & \cellcolor{yellow!60}25.80 & \cellcolor{yellow!60}27.08 & \cellcolor{yellow!60}26.45 \\
 &  & \textit{snow}          & 26.86 & 30.93 & 45.82 & 57.09 & 67.48 & 45.64 \\
 &  & \textit{fog}           & 0.05 & 0.49 & 1.22 & 2.68 & 4.94 & 1.88  \\ \cmidrule(r){2-9} 
 & \multirow{5}{*}{\textbf{Noise}}   & \textit{uniform\_rad}  & 0.46 & 2.48 & 6.27 & 11.64 & 18.22 & 7.81  \\  
 &  & \textit{gaussian\_rad} & 1.65 & 4.36 & 8.81 & 13.67 & 19.74 & 9.65  \\
 &  & \textit{impulse\_rad}  & 1.05 & 1.53 & 1.98 & 2.74 & 4.97 & 2.45  \\
 &  & \textit{background}    & 2.13 & 2.49 & 2.90 & 3.28 & 5.48 & 3.26  \\
 &  & \textit{upsample}      & 0.30 & 0.31 & 0.53 & 0.75 & 1.85 & 0.75  \\ \cmidrule(r){2-9} 
 & \multirow{5}{*}{\textbf{Density}} & \textit{cutout}        & 1.86 & 2.31 & 3.90 & 5.04 & 7.17 & 4.06  \\  
 &  & \textit{local\_dec}    & 5.20 & 6.71 & 9.44 & 15.05 & 35.82 & 14.44 \\
 &  & \textit{local\_inc}    & 0.82 & 1.05 & 1.53 & 2.10 & 2.92 & 1.68  \\
 &  & \textit{beam\_del}     & 0.05 & 0.10 & 0.41 & 0.93 & 2.15 & 0.73  \\
 &  & \textit{layer\_del}    & 0.39 & 2.23 & 2.82 & 4.53 & 5.89 & 3.17  \\ \midrule
\multirow{12}{*}{\rotatebox{90}{\textbf{Object-level}}} & \multirow{4}{*}{\textbf{Noise}} & \textit{uniform} & 0.62 & 2.04 & 5.78 & 12.57 & 23.71 & 8.94 \\  
 &  & \textit{gaussian}      & 1.54 & 4.26 & 9.09 & 17.54 & 29.66 & 12.42 \\
 &  & \textit{impulse}       & 1.86 & 2.42 & 3.08 & 3.64 & 5.31 & 3.26  \\
 &  & \textit{upsample}      & 0.37 & 0.54 & 0.57 & 0.79 & 1.45 & 0.74  \\ \cmidrule(r){2-9} 
 & \multirow{3}{*}{\textbf{Density}} & \textit{cutout}        & 5.97 & 11.45 & 16.08 & 20.29 & 23.56 & 15.47 \\  
 &  & \textit{local\_dec}    & 2.10 & 10.35 & 15.00 & 19.02 & 22.70 & 13.83 \\
 &  & \textit{local\_inc}    & 8.04 & 11.87 & 13.82 & 14.85 & 15.46 & 12.81 \\ \cmidrule(r){2-9} 
 & \multirow{5}{*}{\textbf{Transformation}} & \textit{shear} & 3.99 & 15.49 & 37.54 & 63.85 & 77.67 & 39.71 \\  
 &  & \textit{FFD}           & 2.39 & 14.04 & 34.85 & 55.59 & 67.78 & 34.93 \\
 &  & \textit{rotation}      & 0.05 & 0.19 & 0.26 & 0.84 & 1.25 & 0.52  \\
 &  & \textit{scale}       & 0.46 & 2.34 & 5.23 & 9.65 & 17.16 & 6.97  \\
 &  & \textit{translation}   & 0.98 & 3.85 & 5.06 & 4.22 & 4.75 & 3.77  \\ \midrule
\multicolumn{3}{c|}{\textbf{Average}}  & 3.85 & 6.43 & 10.30 & 14.73 & 19.77 & 11.01 \\ \bottomrule
\end{tabular}%
}
\end{table}

\subsection{Effects of Common Corruptions to Point Cloud Detectors}
\label{sec: main_Effects of Common Corruptions to Point Cloud Detectors}

{\noindent \bf How do different corruptions affect detectors' overall accuracy?}
% over gap after corruptions
% Table~\ref{tab: CEs of different detectors and CCs} shows the AP loss of detectors on Car detection under all common corruptions. 
As shown in the yellow cell in Table~\ref{tab: CEs of different detectors and CCs}, the average $mCE_{AP}$ of $11.01\%$ anticipates a noticeable accuracy drop of detectors against diverse corruption patterns. These results suggest that there is an urgent need of addressing the point cloud detector's robustness issue.

Specifically, \{\textit{rain, snow}\} and \{\textit{shear, FFD}\} corruptions have the AP loss of more than $20\%$ (last column in Table~\ref{tab: CEs of different detectors and CCs}), which presents a serious degradation of detection accuracy. By contrast, some corruption patterns (\eg scene-level and object-level \textit{upsample}, scene-level \textit{beam\_del}, and object-level \textit{rotation}) show less effects on detectors ($CE_{AP}$ less than $1\%$). It demonstrates that upsampling noise, sparse beam loss, and slight rotation don't affect detectors' accuracy. 

Besides, as shown in Table~\ref{tab: CE_Recalls of different detectors and CCs}, the recall metric performs similarly to AP, as the serious recall loss of over $20\%$ happens to \{\textit{rain, snow, shear}\}. In addition, object-level \{\textit{cutout, local\_dec, FFD}\} present an unignorable drop of recall within $[15\%, 20\%]$.

%check the trends of CEs as severity levels up and find exceptions and explain them

{\noindent \bf How do corruption severity levels affect detectors' overall accuracy?} We find almost all common corruptions have a predictable trend, \ie, each corruption's $CE_{AP}$ increases as its severity level increases (see Table~\ref{tab: CEs of different CCs and severities} for detailed results). The only exception is \textit{rain}, $CE_{AP}$ of which remain around $26\%$ regardless of the severity level. There are two plausible explanations: (1) noise points reflected by rain droplets are too sparse to affect detection (see Figure~\ref{fig: rain_comparison}), and (2) a vast amount of points with zero-value reflection intensity in KITTI are not affected by \textit{rain} corruptions at 1-5 severity levels, which cause a fixed accuracy drop on point cloud detection. 

\begin{figure}[ht]
    \centering
    \includegraphics[width=0.95\linewidth]{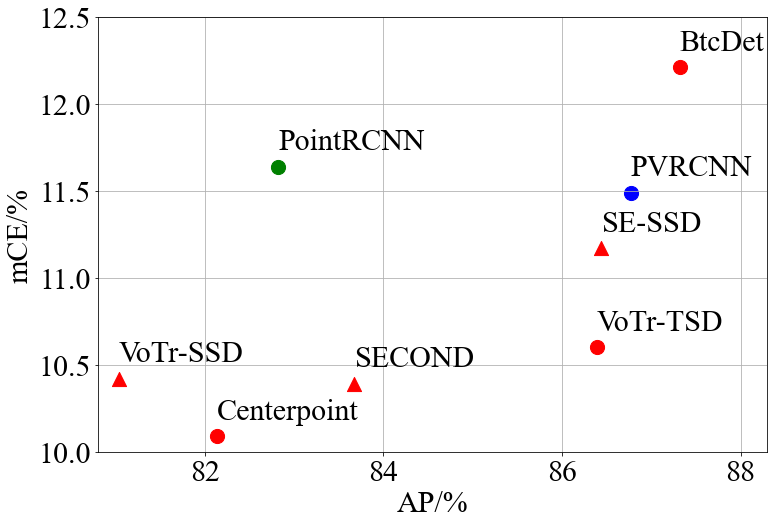}
    \caption{{$mCE_{AP}$ of detectors with different representations on \textit{Car} detection (\{{\color{red}red}, {\color{green} green}, {\color{blue} blue}\} for \{voxel-based, point-based, voxel-point-based\} detectors and \{circle, triangle\} for \{two-stage, one-stage\} ones)}}
    \label{fig:AP_CE}
\end{figure}

\subsection{Reacts of Detector Designing to Common Corruptions}
\label{sec: main_Reacts of Detector Designing to Common Corruptions}

\begin{table}[t]
\centering
\caption{$CE_{AP}(\%)$ of detectors with different proposal architectures on \textit{Car} detection({\colorbox{green!40} {green}} cell for the lower mean CE between one-stage and two-stage detector under a certain corruption)}
\label{tab: CEs of different proposal architectures}
\resizebox{0.45\textwidth}{!}{%
\begin{tabular}{c|c|c|c|c}
\toprule
\multicolumn{3}{c|}{\multirow{1}{*}{\textbf{Corruption}}} & \textbf{one-stage} & \textbf{two-stage} \\  \midrule
\multirow{13}{*}{\rotatebox{90}{\textbf{Scene-level}}} & \multirow{3}{*}{\textbf{Weather}} & \textit{rain} & 26.50 & \cellcolor{green!40}26.42 \\
 &  & \textit{snow} & 46.04 & \cellcolor{green!40}45.39 \\
 &  & \textit{fog} & \cellcolor{green!40}  2.01 \\ \cmidrule(r){2-5} 
 & \multirow{5}{*}{\textbf{Noise}} & \textit{uniform\_rad} & \cellcolor{green!40} 7.55 &  7.98 \\  
 &  & \textit{gaussian\_rad} & \cellcolor{green!40} 9.33 & 9.84 \\
 &  & \textit{impulse\_rad} & \cellcolor{green!40} 1.89  & 2.80 \\
 &  & \textit{background} & \cellcolor{green!40} 3.05 &  3.38 \\
 &  & \textit{upsample} &  \cellcolor{green!40} 0.41 & 0.95 \\ \cmidrule(r){2-5} 
 & \multirow{5}{*}{\textbf{Density}} & \textit{cutout} &  4.35 & \cellcolor{green!40} 3.88 \\  
 &  & \textit{local\_dec} & 15.12 & \cellcolor{green!40} 13.93 \\
 &  & \textit{local\_inc} & \cellcolor{green!40} 1.30  & 1.92 \\
 &  & \textit{beam\_del} & 0.87 & \cellcolor{green!40}0.65 \\
 &  & \textit{layer\_del} & 3.25 &  \cellcolor{green!40} 3.12 \\ \midrule
\multirow{12}{*}{\rotatebox{90}{\textbf{Object-level}}} & \multirow{4}{*}{\textbf{Noise}} & \textit{uniform} & \cellcolor{green!40} 6.41 & 10.46 \\  
 &  & \textit{gaussian} &  \cellcolor{green!40} 9.09 & 14.42 \\
 &  & \textit{impulse} &  \cellcolor{green!40} 2.54 & 3.69 \\
 &  & \textit{upsample} &  \cellcolor{green!40} 0.32 & 0.99 \\ \cmidrule(r){2-5} 
 & \multirow{3}{*}{\textbf{Density}} & \textit{cutout} &  15.52 & \cellcolor{green!40} 15.44 \\  
 &  & \textit{local\_dec} &  \cellcolor{green!40} 13.66  & 13.95 \\
 &  & \textit{local\_inc} &  \cellcolor{green!40} 12.04 &  13.27 \\ \cmidrule(r){2-5} 
 & \multirow{5}{*}{\textbf{Transformation}} & \textit{shear} &  40.07 &  \cellcolor{green!40} 39.49 \\  
 &  & \textit{FFD} &  \cellcolor{green!40} 34.75 &  35.04 \\
 &  & \textit{rotation} &  \cellcolor{green!40} 0.32 & 0.63 \\
 &  & \textit{scale} &  7.34 &  \cellcolor{green!40} 6.74 \\
 &  & \textit{translation} & \cellcolor{green!40} 3.16 &  4.14 \\ \midrule
\multicolumn{3}{c|}{mCE} &  \cellcolor{green!40} 10.66 & 11.21 \\ \bottomrule
\end{tabular}%
}
\end{table}

{\noindent \bf How do different representations affect detectors?}
%
% representation with highest CE
As shown in Figure~\ref{fig:AP_CE}, voxel-based Centerpoint and BtcDet record the lowest and highest $CE_{AP}$. 
For voxel-involving detection (\ie, except for PointRCNN), $mCE_{AP}$ approximately increases as $AP$ increases. It indicates a potential trend that more accurate models trend to become less robust against common corruptions.
% for each corruptions, find most robust (lowest CE) representation
% try to find the partitions of corruption category

We also find that, for the most of corruptions (except \{\textit{shear, FFD, scale}\}), voxel-based methods are generally more robust against corruption patterns (as shown in Table~\ref{tab: CEs of different detectors and CCs}). 
{One plausible explanation is that the spatial quantization of a group of neighbor points by voxelization mitigates the local randomness and the absence of points caused by Noise and Density corruptions.} 
Specifically, for severe corruptions (\eg, $\textit{shear, FFD}$ in the Transformation), the point-voxel-based method PVRCNN is more robust. The point-based PointRCNN doesn't have the most robust performance against any corruption, suggesting potential limitations.

\begin{table}[t]
\centering
\caption{Bug rates ($\%$) of true detection, false classification, false detection, and missing detection of detectors under different corruptions}
\label{tab: BR_cor}
\resizebox{0.45\textwidth}{!}{%
\begin{tabular}{@{}c|c|c|cccc@{}}
\toprule
\multicolumn{3}{c|}{\textbf{Corruption}} & \textbf{TD} & \textbf{FC} & \textbf{FD} & \textbf{MD} \\ \midrule
\multicolumn{3}{c|}{\textit{Clean}} & 43.81 & 0.37 & 9.81 & 46.01 \\ \midrule
\multirow{13}{*}{\rotatebox{90}{\textbf{Scene-level}}} & \multirow{3}{*}{\textbf{Weather}} & \textit{rain} & 42.93 & 1.12 & 17.43 & 38.52 \\
 &  & \textit{snow} & 35.11 & 1.34 & 19.96 & 43.58 \\
 &  & \textit{fog} & 41.99 & 0.46 & 9.91 & 47.64 \\ \cmidrule(l){2-7} 
 & \multirow{5}{*}{\textbf{Noise}} & \textit{uniform\_rad} & 42.86 & 0.95 & 13.79 & 42.40 \\ 
 &  & \textit{gaussian\_rad} & 42.35 & 1.16 & 14.52 & 41.97 \\
 &  & \textit{impulse\_rad} & 43.37 & 0.46 & 11.70 & 44.48 \\
 &  & \textit{background} & 36.99 & 0.30 & 9.22 & 53.49 \\
 &  & \textit{upsample} & 42.46 & 0.35 & 10.00 & 47.19 \\ \cmidrule(l){2-7} 
 & \multirow{5}{*}{\textbf{Density}} & \textit{cutout} & 41.79 & 0.50 & 10.75 & 46.97 \\  
 &  & \textit{local\_dec} & 39.63 & 0.74 & 13.00 & 46.63 \\
 &  & \textit{local\_inc} & 42.94 & 0.40 & 10.30 & 46.36 \\
 &  & \textit{beam\_del} & 43.90 & 0.41 & 10.24 & 45.44 \\
 &  & \textit{layer\_del} & 42.64 & 0.47 & 10.77 & 46.13 \\ \midrule
\multirow{12}{*}{\rotatebox{90}{\textbf{Object-level}}} & \multirow{4}{*}{\textbf{Noise}} & \textit{uniform} & 41.15 & 0.55 & 12.13 & 46.16 \\ 
 &  & \textit{gaussian} & 40.08 & 0.61 & 12.97 & 46.33 \\
 &  & \textit{impulse} & 42.75 & 0.38 & 10.89 & 45.98 \\
 &  & \textit{upsample} & 43.59 & 0.40 & 10.20 & 45.81 \\ \cmidrule(l){2-7} 
 & \multirow{3}{*}{\textbf{Density}} & \textit{cutout} & 36.82 & 0.66 & 11.16 & 51.36 \\  
 &  & \textit{local\_dec} & 37.70 & 0.58 & 10.69 & 51.02 \\
 &  & \textit{local\_inc} & 39.09 & 0.45 & 14.01 & 46.45 \\ \cmidrule(l){2-7} 
 & \multirow{5}{*}{\textbf{Transformation}} & \textit{shear} & 29.11 & 0.47 & 24.44 & 45.99 \\  
 &  & \textit{FFD} & 31.77 & 0.44 & 21.64 & 46.15 \\
 &  & \textit{rotation} & 43.61 & 0.37 & 10.05 & 45.96 \\
 &  & \textit{scaling} & 40.44 & 0.38 & 13.09 & 46.08 \\
 &  & \textit{translation} & 42.08 & 0.47 & 11.45 & 46.00 \\ \bottomrule
\end{tabular}%
}
\end{table}

% \begin{figure*}[hb]
%   \centering
%   \includegraphics[width=1\linewidth]{image/BR_cor.png}
%   \caption{Distribution of bugs under different corruptions}
%   \label{fig: BR_cor}
% \end{figure*}

% find the more robust proposal architecture (lowest CE)
{\noindent \bf How do different proposal architectures affect detectors?} As shown in Figure~\ref{fig:AP_CE}, 
two-stage detectors perform less robust against common corruptions compared to one-stage detectors, showing a lower $mCE_{AP}$. One possible cause is that corrupted data could affect the proposal generation of stage 1 (for two-stage detectors and one-stage ones), and the low-quality proposals significantly affect the BBox regression of stage 2 (only for two-stage detectors).
% for each corruptions, find most robust (lowest CE) proposal architecture
% try to find the partitions of corruption category

\begin{table}[t]
\centering
\caption{Bug rates ($\%$) of true detection, false classification, false detection, and missing detection of different detectors under corruptions (testing results on \textit{clean} data are in parentheses)}
\label{tab: BR_det}
\resizebox{0.48\textwidth}{!}{%
\begin{tabular}{@{}c|cccc@{}}
\toprule
\textbf{Detector} & \multicolumn{1}{c}{\textbf{TD}} & \multicolumn{1}{c}{\textbf{FC}} & \multicolumn{1}{c}{\textbf{FD}} & \multicolumn{1}{c}{\textbf{MD}} \\ \midrule
\textbf{PVRCNN} & 32.63 (36.27) & 1.16 (0.72) & 11.1 (8.28) & 55.11 (54.73) \\
\textbf{PointRCNN} & 47.11 (50.20) & 0.68 (0.40) & 16.04 (12.47) & 36.17 (36.93) \\
\textbf{SECOND} & 20.61 (23.57) & 0.81 (0.49) & 8.42 (6.51) & 70.16 (69.43) \\
\textbf{BtcDet} & 65.24 (68.19) & 0.03 (0.01) & 15.19 (11.13) & 19.54 (20.67) \\
\textbf{VoTr-SSD} & 27.95 (32.17) & 0.62 (0.55) & 13.52 (10.43) & 57.91 (56.85) \\
\textbf{VoTr-TSD} & 42.75 (47.20) & 0.28 (0.20) & 14.67 (10.72) & 42.3 (41.88) \\
\textbf{SE-SSD} & 64.56 (68.18) & 0.05 (0.03) & 15.72 (11.45) & 19.67 (20.34) \\ 
\textbf{Centerpoint} & 21.74 (24.70) & 0.98 (0.54) & 9.26 (7.53) & 68.02 (67.23) \\ \bottomrule
\end{tabular}%
}
\end{table}

As shown in Table \ref{tab: CEs of different proposal architectures}, two-stage detectors present more accurate detection under the scene-level \{\textit{cutout, local\_dec, beam\_del, layer\_del}\} and object-level \{\textit{cutout, shear, scale}\}, displaying a lower average $CE_{AP}$, while one-stage detectors present more accurate under the rest of common corruptions. In summary, one-stage detectors perform more robust against corruptions of scene-level Noise and object-level Noise and Density, while two-stage detectors are mainly more robust against Weather and scene-level Density. 
As for Transformation corruptions, one-stage detectors present better robustness on \{\textit{FFD, rotation, translation}\} and two-stage detectors work better under corruptions of \{\textit{shear, scale}\}.

% \begin{figure*}[]
%   \centering
%   \includegraphics[width=1\linewidth]{image/Average_CE_cor_ProArc.png}
%   \caption{$CE_{AP}$ of one-stage and two-stage detectors on \textit{Car} detection under different corruptions}
%   \label{fig: Average_CE_cor_ProArc}
% \end{figure*}

% \begin{figure*}[ht]
%   \centering
%   \includegraphics[width=1\linewidth]{image/BR_det.png}
%   \caption{Distribution of bugs in different detectors }
%   \label{fig: BR_det}
% \end{figure*}

\subsection{Detection Bugs in Detectors under Common Corruptions}
\label{sec: main_Detection Bugs in Detectors under Common Corruptions}

{\noindent \bf How do different corrupted inputs trigger bugs in detectors?}
We find that the rate of false classification (FC) against common corruption patterns is relatively small, where the largest $CR_{FC}$ is only $0.97\%$ (refer to Table~\ref{tab: FC_BR} in Appendix \ref{sec: Detection Bugs in Detectors under Common Corruptions}). By contrast, the increase of false detection (FD) rate is relatively obvious, by the average $CR_{FD}$ of $3.17\%$ and the largest $CR_{FD}$ of $14.61\%$ (refer to Table~\ref{tab: FD_BR} in Appendix \ref{sec: Detection Bugs in Detectors under Common Corruptions}). 
%Regarding missed detection (MD) (Table~\ref{tab: MD_BR} in Appendix \ref{sec: Detection Bugs in Detectors under Common Corruptions}), \{\textit{rain}\} and scene-level \{\textit{uniform\_rad, gaussian\_rad}\} trigger a drop of MD rate of more than $3\%$, while scene-level \{\textit{background}\} and object-level \{\textit{cutout, local\_dec}\} result in an increase of MD rate of more than $5\%$.
Regarding missed detection (MD) (refer to Table~\ref{tab: MD_BR} in Appendix \ref{sec: Detection Bugs in Detectors under Common Corruptions}), scene-level \{\textit{background}\} and object-level \{\textit{cutout, local\_dec}\} result in an increase of MD rate of more than $5\%$.
% above-found corruptions, find out the change patterns

Surprisingly, according to Table~\ref{tab: BR_cor}, \{\textit{rain}\} and scene-level \{\textit{uniform\_rad, gaussian\_rad}\} even reduce the rate of missing objects. One plausible explanation for this observation is that milder noise points offer a better knowledge of the shape of some objects to detectors, but positioning on those objects is not accurate since the rate of false detection increases (more details in Table \ref{tab: FD_BR} and \ref{tab: MD_BR} in Appendix \ref{sec: Detection Bugs in Detectors under Common Corruptions}).

\begin{figure}[b]
  \centering
  \includegraphics[width=\linewidth]{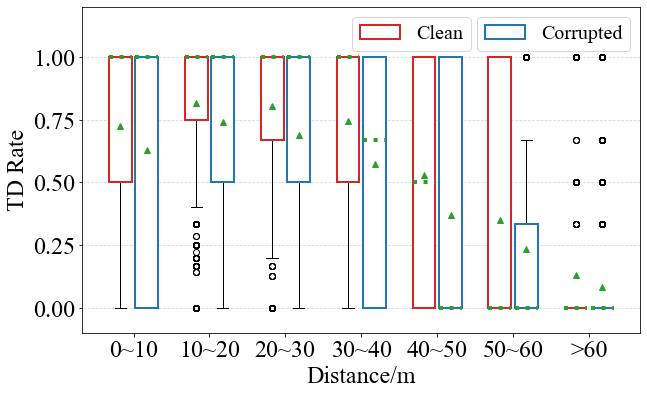}
  \caption{Box-plot of TD rate of all frames \wrt different distances of objects to the LiDAR sensor (green dotted lines for the median and green triangles for the mean)}
  \label{fig: TD_dis}
\end{figure}

Also, we find that, as shown in Figure \ref{fig: TD_dis}, compared to clean observations, TD rates under corrupted observations are always lower at any distance of objects to LiDAR.

%(refer to Figures \ref{fig: TD_dis_scene} and \ref{fig: TD_dis_object} in Appendix \ref{sec: Detection Bugs in Detectors under Common Corruptions}). 

{\noindent \bf How do corrupted inputs trigger bugs in different detectors?}
%
% find the overall changes of FC, FD, MD   
%
% above-found detectors, find out the change patterns
In general, as shown in Table \ref{tab: BR_det}, most of the detectors perform relatively stable in terms of false classification rates and missed detection rates against common corruptions. In contrast, affected by corruptions, all detectors have increasing false detection rates (refer to Table \ref{tab: BR_det}), revealing a serious bias in BBox localization of point cloud detection. Among all detectors, BtcDet and SE-SSD records a serious FD increase of over $4\%$.

\begin{table*}[ht]
\centering
\caption{Average $CE_{AP}(\%)$ of detectors given different common corruptions on \textit{Car} detection with DA and/or denoising (the differences between enhancement methods and \textbf{Origin} are in parentheses)}
\label{tab: CE given different common corruptions with DA and/or Denoising}
\resizebox{0.7\textwidth}{!}{%
\begin{tabular}{c|c|c|cccc}
\toprule
\multicolumn{3}{c|}{\textbf{Corruption}} & \textbf{Origin} & \textbf{PA-DA} & \textbf{KNN-RO} & \textbf{PA-DA + KNN-RO} \\ \midrule
\multirow{13}{*}{\textbf{Scene-level}} & \multirow{3}{*}{\textbf{Weather}} & \textit{rain} & 26.45 & 27.51 (+1.06) & 32.03 (+5.58) & 32.79 (+6.34) \\
 &  & \textit{snow} & 45.64 & 45.68 (+0.04) & 47.76 (+2.12) & 47.83 (+2.19) \\
 &  & \textit{fog} & 1.88 & 2.0 (+0.12) & 5.18 (+3.3) & 5.09 (+3.21) \\ \cmidrule (r){2-7} 
 & \multirow{5}{*}{\textbf{Noise}} & \textit{uniform\_rad} & 7.82 & 7.85 (+0.03) & 10.79 (+2.97) & 10.79 (+2.97) \\
 &  & \textit{gaussian\_rad} & 9.65 & 9.59 (-0.06) & 12.65 (+3.0) & 12.47 (+2.82) \\
 &  & \textit{impulse\_rad} & 2.46 & 2.01 (-0.45) & 4.99 (+2.53) & 4.53 (+2.07) \\
 &  & \textit{background} & 3.25 & 3.08 (-0.17) & 2.36 (-0.89) & 2.13 (-1.12) \\
 &  & \textit{upsample} & 0.75 & 0.65 (-0.1) & 3.61 (+2.86) & 3.4 (+2.65) \\ \cmidrule (r){2-7} 
 & \multirow{5}{*}{\textbf{Density}} & \textit{cutout} & 4.06 & 3.97 (-0.09) & 6.71 (+2.65) & 6.52 (+2.46) \\
 &  & \textit{local\_dec} & 14.44 & 14.83 (+0.39) & 16.73 (+2.29) & 17.0 (+2.56) \\
 &  & \textit{local\_inc} & 1.68 & 1.49 (-0.19) & 4.39 (+2.71) & 4.21 (+2.53) \\
 &  & \textit{beam\_del} & 0.73 & 0.77 (+0.04) & 3.45 (+2.72) & 3.39 (+2.66) \\
 &  & \textit{layer\_del} & 3.17 & 3.19 (+0.02) & 6.23 (+3.06) & 6.2 (+3.03) \\ \midrule
\multirow{12}{*}{\textbf{Object-level}} & \multirow{4}{*}{\textbf{Noise}} & \textit{uniform} & 8.94 & 7.95 (-0.99) & 11.55 (+2.61) & 10.66 (+1.72) \\
 &  & \textit{gaussian} & 12.42 & 11.46 (-0.96) & 14.99 (+2.57) & 14.05 (+1.63) \\
 &  & \textit{impulse} & 3.26 & 2.96 (-0.3) & 6.07 (+2.81) & 5.76 (+2.5) \\
 &  & \textit{upsample} & 0.74 & 0.57 (-0.17) & 2.77 (+2.03) & 2.4 (+1.66) \\ \cmidrule (r){2-7} 
 & \multirow{3}{*}{\textbf{Density}} & \textit{cutout} & 15.47 & 15.1 (-0.37) & 17.15 (+1.68) & 16.8 (+1.33) \\
 &  & \textit{local\_dec} & 13.84 & 13.62 (-0.22) & 16.18 (+2.34) & 15.92 (+2.08) \\
 &  & \textit{local\_inc} & 12.81 & 11.8 (-1.01) & 15.01 (+2.2) & 14.14 (+1.33) \\ \cmidrule (r){2-7} 
 & \multirow{5}{*}{\textbf{Transformation}} & \textit{shear} & 39.71 & 39.72 (+0.01) & 40.06 (+0.35) & 39.97 (+0.26) \\
 &  & \textit{FFD} & 34.93 & 34.34 (-0.59) & 40.55 (+5.62) & 40.02 (+5.09) \\
 &  & \textit{rotation} & 0.52 & 0.52 (+0.0) & 3.19 (+2.67) & 3.19 (+2.67) \\
 &  & \textit{scaling} & 6.97 & 7.01 (+0.04) & 9.4 (+2.43) & 9.42 (+2.45) \\
 &  & \textit{translation} & 3.77 & 3.48 (-0.29) & 6.22 (+2.45) & 5.76 (+1.99) \\ \midrule
\multicolumn{3}{c|}{Average} & 11.01 & 10.85 (-0.16) & 13.6 (+2.59) & 13.38 (+2.37) \\ \bottomrule
\end{tabular}
}
\end{table*}

\subsection{Robustness Enhancement by Data Augmentation and Denoising}
\label{sec: main_Robustness Enhancement by Data Augmentation and Denoising}

{\noindent \bf How do PA-DA and KNN-based outlier-removing affect detectors' robustness against different corruptions?}
% find overall CE changes of DA, DE, and DA-DE 
% respectively for DA, DE, and DA-DE, find the corruptions with CE down or robustness up 
% PA-DA
Shown by Table \ref{tab: CE given different common corruptions with DA and/or Denoising}, the average $CE_{AP}$ with PA-DA slightly decreased to $10.85\%$ compared to the average $CE_{AP}$ without PA-DA, which still poses serious robustness issues for point cloud detectors. 
% By adopting PA-DA, the AP on \textit{Car} detection against object-level \{\textit{uniform, gaussian, local\_inc}.
% KNN-RO

Regarding denoising strategy, the average $CE_{AP}$ after adopting KNN-RO increases to $13.60\%$ without PA-DA and $13.38\%$ with PA-DA (refer to Table \ref{tab: CE given different common corruptions with DA and/or Denoising}). These results indicate that KNN-RO might not be capable of enhancing point cloud detectors' robustness against most of the corruptions in \textit{Car} detection. However, we find that KNN-RO slightly improves the robustness of \textit{Pedestrian} detection by decreasing the $CE_{AP}$ by $0.14\%$ without PA-DA and $1.19\%$ with PA-DA (Table~\ref{tab: CE given different common corruptions with DA and/or Denoising on pedestrian detection} in Appendix \ref{sec: Robustness Enhancement by Data Augmentation and Denoising}).

\begin{figure*}[ht]
  \centering
  \includegraphics[width=1\linewidth]{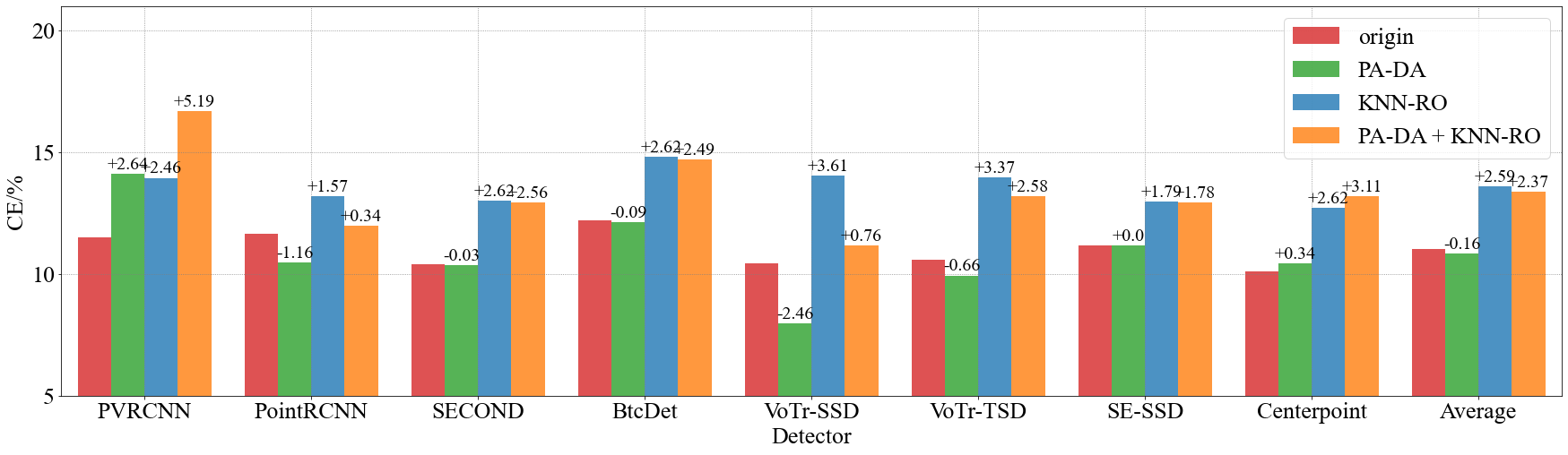}
  \caption{Average $CE_{AP}(\%)$ of different detectors on \textit{Car} detection given common corruptions }
  \label{fig: Average_CE_DA&DE_dets}
\end{figure*}

{\noindent \bf How do PA-DA and KNN-based outlier-removing affect different detectors' robustness against corruptions?}
%
% respectively for DA, DE, and DA&DE, find the detectors with CE down or robustness up 
% PA-DA
% As shown in Figure~\ref{fig: Average_CE_DA&DE_dets} in Appendix \ref{sec: Robustness Enhancement by Data Augmentation and Denoising}, except PVRCNN and SE-SSD, the rest detectors present decreasing $CE_{AP}$. Further, PointRCNN and VoTr-SSD obtain the AP increase of $1.16\%$ and $2.46\%$, respectively.
Except for PVRCNN, SE-SSD, and Centerpoint, all the other detectors perform more robust against corruption patterns after adopting PA-DA (refer to Figure~\ref{fig: Average_CE_DA&DE_dets}). Moreover, PointRCNN and VoTr-SSD increase their AP by $1.16\%$ and $2.46\%$ after adopting PA-DA, respectively.
% KNN-RO

According to Figure~\ref{fig: Average_CE_DA&DE_dets}, KNN-RO degrade AP metric for all detectors, presenting no improvement on the robustness of any detector on \textit{Car} detection. However, adopting KNN-RO slightly improves the AP by $0.14\%$ without PA-DA and $1.19\%$ with PA-DA on \textit{Pedestrian} detection, respectively (refer to Table \ref{tab: CE given different detectors with DA and/or Denoising on pedestrian detection} in Appendix \ref{sec: Robustness Enhancement by Data Augmentation and Denoising}). It illustrates that compared with effects on \textit{Car} objects, KNN-RO are more effective in removing perturbations caused by corruptions for \textit{Pedestrian} objects.

\section{Conclusion}

In this paper, we propose the first physical-aware robustness benchmark of point cloud detection against common corruption patterns, {which contains a total of 1,122,150 examples covering 25 common corruption types and 6 severity levels.} Based on the benchmark, we conduct extensive empirical studies on {8 detectors covering 6 different detection frameworks and reveal the vulnerabilities of the detectors.} 
{Moreover, we further study the effectiveness of existing data augmentation and denoising methods and find them limited, calling for more research on robustness enhancement.} We hope this benchmark and empirical study results can guide future research toward building more robust and reliable point cloud detectors.

% \begin{ack}

% \end{ack}

\bibliographystyle{IEEEtran}
\bibliography{ref}

\begin{thebibliography}{10}
\providecommand{\url}[1]{#1}
\csname url@samestyle\endcsname
\providecommand{\newblock}{\relax}
\providecommand{\bibinfo}[2]{#2}
\providecommand{\BIBentrySTDinterwordspacing}{\spaceskip=0pt\relax}
\providecommand{\BIBentryALTinterwordstretchfactor}{4}
\providecommand{\BIBentryALTinterwordspacing}{\spaceskip=\fontdimen2\font plus
\BIBentryALTinterwordstretchfactor\fontdimen3\font minus
  \fontdimen4\font\relax}
\providecommand{\BIBforeignlanguage}[2]{{%
\expandafter\ifx\csname l@#1\endcsname\relax
\typeout{** WARNING: IEEEtran.bst: No hyphenation pattern has been}%
\typeout{** loaded for the language `#1'. Using the pattern for}%
\typeout{** the default language instead.}%
\else
\language=\csname l@#1\endcsname
\fi
#2}}
\providecommand{\BIBdecl}{\relax}
\BIBdecl

\bibitem{guo2020deep}
Y.~Guo, H.~Wang, Q.~Hu, H.~Liu, L.~Liu, and M.~Bennamoun, ``Deep learning for
  3d point clouds: A survey,'' \emph{IEEE transactions on pattern analysis and
  machine intelligence}, vol.~43, no.~12, pp. 4338--4364, 2020.

\bibitem{qian20213d}
R.~Qian, X.~Lai, and X.~Li, ``3d object detection for autonomous driving: a
  survey,'' \emph{arXiv preprint arXiv:2106.10823}, 2021.

\bibitem{badue2021self}
C.~Badue, R.~Guidolini, R.~V. Carneiro, P.~Azevedo, V.~B. Cardoso, A.~Forechi,
  L.~Jesus, R.~Berriel, T.~M. Paixao, F.~Mutz \emph{et~al.}, ``Self-driving
  cars: A survey,'' \emph{Expert Systems with Applications}, vol. 165, p.
  113816, 2021.

\bibitem{arnold2019survey}
E.~Arnold, O.~Y. Al-Jarrah, M.~Dianati, S.~Fallah, D.~Oxtoby, and
  A.~Mouzakitis, ``A survey on 3d object detection methods for autonomous
  driving applications,'' \emph{IEEE Transactions on Intelligent Transportation
  Systems}, vol.~20, no.~10, pp. 3782--3795, 2019.

\bibitem{fernandes2021point}
D.~Fernandes, A.~Silva, R.~N{\'e}voa, C.~Simoes, D.~Gonzalez, M.~Guevara,
  P.~Novais, J.~Monteiro, and P.~Melo-Pinto, ``Point-cloud based 3d object
  detection and classification methods for self-driving applications: A survey
  and taxonomy,'' \emph{Information Fusion}, vol.~68, pp. 161--191, 2021.

\bibitem{geiger2013vision}
A.~Geiger, P.~Lenz, C.~Stiller, and R.~Urtasun, ``Vision meets robotics: The
  kitti dataset,'' \emph{The International Journal of Robotics Research},
  vol.~32, no.~11, pp. 1231--1237, 2013.

\bibitem{caesar2020nuscenes}
H.~Caesar, V.~Bankiti, A.~H. Lang, S.~Vora, V.~E. Liong, Q.~Xu, A.~Krishnan,
  Y.~Pan, G.~Baldan, and O.~Beijbom, ``nuscenes: A multimodal dataset for
  autonomous driving,'' in \emph{Proceedings of the IEEE/CVF conference on
  computer vision and pattern recognition}, 2020, pp. 11\,621--11\,631.

\bibitem{sun2020scalability}
P.~Sun, H.~Kretzschmar, X.~Dotiwalla, A.~Chouard, V.~Patnaik, P.~Tsui, J.~Guo,
  Y.~Zhou, Y.~Chai, B.~Caine \emph{et~al.}, ``Scalability in perception for
  autonomous driving: Waymo open dataset,'' in \emph{Proceedings of the
  IEEE/CVF conference on computer vision and pattern recognition}, 2020, pp.
  2446--2454.

\bibitem{rasshofer2011influences}
R.~H. Rasshofer, M.~Spies, and H.~Spies, ``Influences of weather phenomena on
  automotive laser radar systems,'' \emph{Advances in radio science}, vol.~9,
  no. B. 2, pp. 49--60, 2011.

\bibitem{Bijelic_2020_STF}
M.~Bijelic, T.~Gruber, F.~Mannan, F.~Kraus, W.~Ritter, K.~Dietmayer, and
  F.~Heide, ``Seeing through fog without seeing fog: Deep multimodal sensor
  fusion in unseen adverse weather,'' in \emph{The IEEE Conference on Computer
  Vision and Pattern Recognition (CVPR)}, June 2020.

\bibitem{sun2022benchmarking}
J.~Sun, Q.~Zhang, B.~Kailkhura, Z.~Yu, C.~Xiao, and Z.~M. Mao, ``Benchmarking
  robustness of 3d point cloud recognition against common corruptions,''
  \emph{arXiv preprint arXiv:2201.12296}, 2022.

\bibitem{kilic2021lidar}
V.~Kilic, D.~Hegde, V.~Sindagi, A.~B. Cooper, M.~A. Foster, and V.~M. Patel,
  ``Lidar light scattering augmentation (lisa): Physics-based simulation of
  adverse weather conditions for 3d object detection,'' \emph{arXiv preprint
  arXiv:2107.07004}, 2021.

\bibitem{xu2021behind}
Q.~Xu, Y.~Zhong, and U.~Neumann, ``Behind the curtain: Learning occluded shapes
  for 3d object detection,'' \emph{arXiv preprint arXiv:2112.02205}, 2021.

\bibitem{pitropov2021canadian}
M.~Pitropov, D.~E. Garcia, J.~Rebello, M.~Smart, C.~Wang, K.~Czarnecki, and
  S.~Waslander, ``Canadian adverse driving conditions dataset,'' \emph{The
  International Journal of Robotics Research}, vol.~40, no. 4-5, pp. 681--690,
  2021.

\bibitem{burnett2022boreas}
K.~Burnett, D.~J. Yoon, Y.~Wu, A.~Z. Li, H.~Zhang, S.~Lu, J.~Qian, W.-K. Tseng,
  A.~Lambert, K.~Y. Leung \emph{et~al.}, ``Boreas: A multi-season autonomous
  driving dataset,'' \emph{arXiv preprint arXiv:2203.10168}, 2022.

\bibitem{hendrycks2019benchmarking}
D.~Hendrycks and T.~Dietterich, ``Benchmarking neural network robustness to
  common corruptions and perturbations,'' \emph{arXiv preprint
  arXiv:1903.12261}, 2019.

\bibitem{barbu2019objectnet}
A.~Barbu, D.~Mayo, J.~Alverio, W.~Luo, C.~Wang, D.~Gutfreund, J.~Tenenbaum, and
  B.~Katz, ``Objectnet: A large-scale bias-controlled dataset for pushing the
  limits of object recognition models,'' \emph{Advances in neural information
  processing systems}, vol.~32, 2019.

\bibitem{ren2022benchmarking}
J.~Ren, L.~Pan, and Z.~Liu, ``Benchmarking and analyzing point cloud
  classification under corruptions,'' \emph{arXiv preprint arXiv:2202.03377},
  2022.

\bibitem{chang2019argoverse}
M.-F. Chang, J.~Lambert, P.~Sangkloy, J.~Singh, S.~Bak, A.~Hartnett, D.~Wang,
  P.~Carr, S.~Lucey, D.~Ramanan \emph{et~al.}, ``Argoverse: 3d tracking and
  forecasting with rich maps,'' in \emph{Proceedings of the IEEE/CVF Conference
  on Computer Vision and Pattern Recognition}, 2019, pp. 8748--8757.

\bibitem{Woven@2019}
R.~Kesten, M.~Usman, J.~Houston, T.~Pandya, K.~Nadhamuni, A.~Ferreira, M.~Yuan,
  B.~Low, A.~Jain, P.~Ondruska, S.~Omari, S.~Shah, A.~Kulkarni, A.~Kazakova,
  C.~Tao, L.~Platinsky, W.~Jiang, and V.~Shet, ``Level 5 perception dataset
  2020,'' \url{https://level-5.global/level5/data/}, 2019.

\bibitem{bolkas2018effect}
D.~Bolkas and A.~Martinez, ``Effect of target color and scanning geometry on
  terrestrial lidar point-cloud noise and plane fitting,'' \emph{Journal of
  applied geodesy}, vol.~12, no.~1, pp. 109--127, 2018.

\bibitem{villa2021spads}
F.~Villa, F.~Severini, F.~Madonini, and F.~Zappa, ``Spads and sipms arrays for
  long-range high-speed light detection and ranging (lidar),'' \emph{Sensors},
  vol.~21, no.~11, p. 3839, 2021.

\bibitem{Texas2016lidar}
T.~Instruments, ``Lidar pulsed time of flight reference design,'' 2016.

\bibitem{ma2012analysis}
H.~Ma and J.~Wu, ``Analysis of positioning errors caused by platform vibration
  of airborne lidar system,'' in \emph{2012 8th IEEE International Symposium on
  Instrumentation and Control Technology (ISICT) Proceedings}.\hskip 1em plus
  0.5em minus 0.4em\relax IEEE, 2012, pp. 257--261.

\bibitem{wang2021simultaneous}
R.~Wang, B.~Wang, M.~Xiang, C.~Li, S.~Wang, and C.~Song, ``Simultaneous
  time-varying vibration and nonlinearity compensation for one-period
  triangular-fmcw lidar signal,'' \emph{Remote Sensing}, vol.~13, no.~9, p.
  1731, 2021.

\bibitem{mona2012lidar}
L.~Mona, Z.~Liu, D.~M{\"u}ller, A.~Omar, A.~Papayannis, G.~Pappalardo,
  N.~Sugimoto, and M.~Vaughan, ``Lidar measurements for desert dust
  characterization: an overview,'' \emph{Advances in Meteorology}, vol. 2012,
  2012.

\bibitem{morin2021simulated}
X.~Morin-Duchesne and M.~S. Langer, ``Simulated lidar repositioning: a novel
  point cloud data augmentation method,'' \emph{arXiv preprint
  arXiv:2111.10650}, 2021.

\bibitem{wong2020efficient}
C.-C. Wong and C.-M. Vong, ``Efficient outdoor 3d point cloud semantic
  segmentation for critical road objects and distributed contexts,'' in
  \emph{European Conference on Computer Vision}.\hskip 1em plus 0.5em minus
  0.4em\relax Springer, 2020, pp. 499--514.

\bibitem{wang2020train}
Y.~Wang, X.~Chen, Y.~You, L.~E. Li, B.~Hariharan, M.~Campbell, K.~Q.
  Weinberger, and W.-L. Chao, ``Train in germany, test in the usa: Making 3d
  object detectors generalize,'' in \emph{Proceedings of the IEEE/CVF
  Conference on Computer Vision and Pattern Recognition}, 2020, pp.
  11\,713--11\,723.

\bibitem{xiang2019generating}
C.~Xiang, C.~R. Qi, and B.~Li, ``Generating 3d adversarial point clouds,'' in
  \emph{Proceedings of the IEEE/CVF Conference on Computer Vision and Pattern
  Recognition}, 2019, pp. 9136--9144.

\bibitem{liu2019extending}
D.~Liu, R.~Yu, and H.~Su, ``Extending adversarial attacks and defenses to deep
  3d point cloud classifiers,'' in \emph{2019 IEEE International Conference on
  Image Processing (ICIP)}.\hskip 1em plus 0.5em minus 0.4em\relax IEEE, 2019,
  pp. 2279--2283.

\bibitem{abdelfattah2021adversarial}
M.~Abdelfattah, K.~Yuan, Z.~J. Wang, and R.~Ward, ``Adversarial attacks on
  camera-lidar models for 3d car detection,'' in \emph{2021 IEEE/RSJ
  International Conference on Intelligent Robots and Systems (IROS)}.\hskip 1em
  plus 0.5em minus 0.4em\relax IEEE, 2021, pp. 2189--2194.

\bibitem{zhu2021adversarial}
Y.~Zhu, C.~Miao, F.~Hajiaghajani, M.~Huai, L.~Su, and C.~Qiao, ``Adversarial
  attacks against lidar semantic segmentation in autonomous driving,'' in
  \emph{Proceedings of the 19th ACM Conference on Embedded Networked Sensor
  Systems}, 2021, pp. 329--342.

\bibitem{li2016vehicle}
B.~Li, T.~Zhang, and T.~Xia, ``Vehicle detection from 3d lidar using fully
  convolutional network,'' \emph{arXiv preprint arXiv:1608.07916}, 2016.

\bibitem{yang2018pixor}
B.~Yang, W.~Luo, and R.~Urtasun, ``Pixor: Real-time 3d object detection from
  point clouds,'' in \emph{Proceedings of the IEEE conference on Computer
  Vision and Pattern Recognition}, 2018, pp. 7652--7660.

\bibitem{yan2018second}
Y.~Yan, Y.~Mao, and B.~Li, ``Second: Sparsely embedded convolutional
  detection,'' \emph{Sensors}, vol.~18, no.~10, p. 3337, 2018.

\bibitem{mao2021voxel}
J.~Mao, Y.~Xue, M.~Niu, H.~Bai, J.~Feng, X.~Liang, H.~Xu, and C.~Xu, ``Voxel
  transformer for 3d object detection,'' in \emph{Proceedings of the IEEE/CVF
  International Conference on Computer Vision}, 2021, pp. 3164--3173.

\bibitem{shi2019pointrcnn}
S.~Shi, X.~Wang, and H.~Li, ``Pointrcnn: 3d object proposal generation and
  detection from point cloud,'' in \emph{Proceedings of the IEEE/CVF conference
  on computer vision and pattern recognition}, 2019, pp. 770--779.

\bibitem{yang20203dssd}
Z.~Yang, Y.~Sun, S.~Liu, and J.~Jia, ``3dssd: Point-based 3d single stage
  object detector,'' in \emph{Proceedings of the IEEE/CVF conference on
  computer vision and pattern recognition}, 2020, pp. 11\,040--11\,048.

\bibitem{shi2020pv}
S.~Shi, C.~Guo, L.~Jiang, Z.~Wang, J.~Shi, X.~Wang, and H.~Li, ``Pv-rcnn:
  Point-voxel feature set abstraction for 3d object detection,'' in
  \emph{Proceedings of the IEEE/CVF Conference on Computer Vision and Pattern
  Recognition}, 2020, pp. 10\,529--10\,538.

\bibitem{he2020structure}
C.~He, H.~Zeng, J.~Huang, X.-S. Hua, and L.~Zhang, ``Structure aware
  single-stage 3d object detection from point cloud,'' in \emph{Proceedings of
  the IEEE/CVF Conference on Computer Vision and Pattern Recognition}, 2020,
  pp. 11\,873--11\,882.

\bibitem{deng2009imagenet}
J.~Deng, W.~Dong, R.~Socher, L.-J. Li, K.~Li, and L.~Fei-Fei, ``Imagenet: A
  large-scale hierarchical image database,'' in \emph{2009 IEEE conference on
  computer vision and pattern recognition}.\hskip 1em plus 0.5em minus
  0.4em\relax Ieee, 2009, pp. 248--255.

\bibitem{wu20153d}
Z.~Wu, S.~Song, A.~Khosla, F.~Yu, L.~Zhang, X.~Tang, and J.~Xiao, ``3d
  shapenets: A deep representation for volumetric shapes,'' in
  \emph{Proceedings of the IEEE conference on computer vision and pattern
  recognition}, 2015, pp. 1912--1920.

\bibitem{zhang2021pointcutmix}
J.~Zhang, L.~Chen, B.~Ouyang, B.~Liu, J.~Zhu, Y.~Chen, Y.~Meng, and D.~Wu,
  ``Pointcutmix: Regularization strategy for point cloud classification,''
  \emph{arXiv preprint arXiv:2101.01461}, 2021.

\bibitem{lee2021regularization}
D.~Lee, J.~Lee, J.~Lee, H.~Lee, M.~Lee, S.~Woo, and S.~Lee, ``Regularization
  strategy for point cloud via rigidly mixed sample,'' in \emph{Proceedings of
  the IEEE/CVF Conference on Computer Vision and Pattern Recognition}, 2021,
  pp. 15\,900--15\,909.

\bibitem{choi2020part}
J.~Choi, Y.~Song, and N.~Kwak, ``Part-aware data augmentation for 3d object
  detection in point cloud,'' in \emph{2021 IEEE/RSJ International Conference
  on Intelligent Robots and Systems (IROS)}.\hskip 1em plus 0.5em minus
  0.4em\relax IEEE, 2020, pp. 3391--3397.

\bibitem{fang2021lidar}
J.~Fang, X.~Zuo, D.~Zhou, S.~Jin, S.~Wang, and L.~Zhang, ``Lidar-aug: A general
  rendering-based augmentation framework for 3d object detection,'' in
  \emph{Proceedings of the IEEE/CVF Conference on Computer Vision and Pattern
  Recognition}, 2021, pp. 4710--4720.

\bibitem{hahner2021fog}
M.~Hahner, C.~Sakaridis, D.~Dai, and L.~Van~Gool, ``Fog simulation on real
  lidar point clouds for 3d object detection in adverse weather,'' in
  \emph{Proceedings of the IEEE/CVF International Conference on Computer
  Vision}, 2021, pp. 15\,283--15\,292.

\bibitem{yu2021point}
X.~Yu, L.~Tang, Y.~Rao, T.~Huang, J.~Zhou, and J.~Lu, ``Point-bert:
  Pre-training 3d point cloud transformers with masked point modeling,''
  \emph{arXiv preprint arXiv:2111.14819}, 2021.

\bibitem{zhang2021self}
Z.~Zhang, R.~Girdhar, A.~Joulin, and I.~Misra, ``Self-supervised pretraining of
  3d features on any point-cloud,'' in \emph{Proceedings of the IEEE/CVF
  International Conference on Computer Vision}, 2021, pp. 10\,252--10\,263.

\bibitem{duan2021low}
Y.~Duan, C.~Yang, H.~Chen, W.~Yan, and H.~Li, ``Low-complexity point cloud
  denoising for lidar by pca-based dimension reduction,'' \emph{Optics
  Communications}, vol. 482, p. 126567, 2021.

\bibitem{ning2018efficient}
X.~Ning, F.~Li, G.~Tian, and Y.~Wang, ``An efficient outlier removal method for
  scattered point cloud data,'' \emph{PloS one}, vol.~13, no.~8, p. e0201280,
  2018.

\bibitem{carrilho2018statistical}
A.~Carrilho, M.~Galo, and R.~Santos, ``Statistical outlier detection method for
  airborne lidar data.'' \emph{International Archives of the Photogrammetry,
  Remote Sensing \& Spatial Information Sciences}, 2018.

\bibitem{yin2021center}
T.~Yin, X.~Zhou, and P.~Krahenbuhl, ``Center-based 3d object detection and
  tracking,'' in \emph{Proceedings of the IEEE/CVF conference on computer
  vision and pattern recognition}, 2021, pp. 11\,784--11\,793.

\bibitem{ye2020hvnet}
M.~Ye, S.~Xu, and T.~Cao, ``Hvnet: Hybrid voxel network for lidar based 3d
  object detection,'' in \emph{Proceedings of the IEEE/CVF conference on
  computer vision and pattern recognition}, 2020, pp. 1631--1640.

\bibitem{qi2017pointnet}
C.~R. Qi, H.~Su, K.~Mo, and L.~J. Guibas, ``Pointnet: Deep learning on point
  sets for 3d classification and segmentation,'' in \emph{Proceedings of the
  IEEE conference on computer vision and pattern recognition}, 2017, pp.
  652--660.

\bibitem{qi2017pointnet++}
C.~R. Qi, L.~Yi, H.~Su, and L.~J. Guibas, ``Pointnet++: Deep hierarchical
  feature learning on point sets in a metric space,'' \emph{Advances in neural
  information processing systems}, vol.~30, 2017.

\bibitem{van2001art}
D.~A. Van~Dyk and X.-L. Meng, ``The art of data augmentation,'' \emph{Journal
  of Computational and Graphical Statistics}, vol.~10, no.~1, pp. 1--50, 2001.

\bibitem{chen2020pointmixup}
Y.~Chen, V.~T. Hu, E.~Gavves, T.~Mensink, P.~Mettes, P.~Yang, and C.~G. Snoek,
  ``Pointmixup: Augmentation for point clouds,'' in \emph{European Conference
  on Computer Vision}.\hskip 1em plus 0.5em minus 0.4em\relax Springer, 2020,
  pp. 330--345.

\bibitem{cheng2020improving}
S.~Cheng, Z.~Leng, E.~D. Cubuk, B.~Zoph, C.~Bai, J.~Ngiam, Y.~Song, B.~Caine,
  V.~Vasudevan, C.~Li \emph{et~al.}, ``Improving 3d object detection through
  progressive population based augmentation,'' in \emph{European Conference on
  Computer Vision}.\hskip 1em plus 0.5em minus 0.4em\relax Springer, 2020, pp.
  279--294.

\bibitem{van2008visualizing}
L.~Van~der Maaten and G.~Hinton, ``Visualizing data using t-sne.''
  \emph{Journal of machine learning research}, vol.~9, no.~11, 2008.

\bibitem{gretton2006kernel}
A.~Gretton, K.~Borgwardt, M.~Rasch, B.~Sch{\"o}lkopf, and A.~Smola, ``A kernel
  method for the two-sample-problem,'' \emph{Advances in neural information
  processing systems}, vol.~19, 2006.

\bibitem{chen20003d}
B.~Chen and A.~Kaufman, ``3d volume rotation using shear transformations,''
  \emph{Graphical Models}, vol.~62, no.~4, pp. 308--322, 2000.

\bibitem{sederberg1986free}
T.~W. Sederberg and S.~R. Parry, ``Free-form deformation of solid geometric
  models,'' in \emph{Proceedings of the 13th annual conference on Computer
  graphics and interactive techniques}, 1986, pp. 151--160.

\bibitem{zheng2021se}
W.~Zheng, W.~Tang, L.~Jiang, and C.-W. Fu, ``Se-ssd: Self-ensembling
  single-stage object detector from point cloud,'' in \emph{Proceedings of the
  IEEE/CVF Conference on Computer Vision and Pattern Recognition}, 2021, pp.
  14\,494--14\,503.

\bibitem{openpcdet2020}
O.~D. Team, ``Openpcdet: An open-source toolbox for 3d object detection from
  point clouds,'' \url{https://github.com/open-mmlab/OpenPCDet}, 2020.

\bibitem{zhu2019class}
B.~Zhu, Z.~Jiang, X.~Zhou, Z.~Li, and G.~Yu, ``Class-balanced grouping and
  sampling for point cloud 3d object detection,'' \emph{arXiv preprint
  arXiv:1908.09492}, 2019.

\bibitem{enwiki:1102188829}
{Wikipedia contributors}, ``Rain --- {Wikipedia}{,} the free encyclopedia,''
  \url{https://en.wikipedia.org/w/index.php?title=Rain&oldid=1102188829}, 2022,
  [Online; accessed 5-August-2022].

\end{thebibliography}

\clearpage
\begin{appendices}

\section{Details of Empirical Study}
\label{sec: appendix_empirical_study}

Due to the page limit of supplementary materials, we present some tables and figures on our supplementary website \href{https://sites.google.com/ualberta.ca/robustness1pc2detector/}{https://sites.google.com/ualberta.ca/robustness1pc2detector/}. We also encourage readers to refer to this \textbf{supplementary} website for additional details.

% We present some tables and figures on the URL \href{https://sites.google.com/ualberta.ca/robustness1pc2detector/}{https://sites.google.com/ualberta.ca/robustness1pc2detector/} as a "\textbf{supplementary}" for this paper. We encourage readers to refer to the \textbf{supplementary} for more details.

\begin{table}[ht]
\centering
\caption{{Code sources of detectors}}
\label{tab: det_open}
\resizebox{0.5\textwidth}{!}{%
\scriptsize
\begin{tabular}{c|c|l}
\toprule
\textbf{Detector} & \textbf{Platform} & \multicolumn{1}{c}{\textbf{URL}(http://)} \\ \midrule
SECOND & Openpcdet \cite{openpcdet2020}  & github.com/open-mmlab/OpenPCDet \\
PointRCNN & Openpcdet  & github.com/open-mmlab/OpenPCDet \\
PVRCNN & Openpcdet & github.com/open-mmlab/OpenPCDet \\
BtcDet & Openpcdet &  github.com/xharlie/btcdet\\
VoTr-SSD & Openpcdet &  github.com/PointsCoder/VOTR \\
VoTr-TSD & Openpcdet &  github.com/PointsCoder/VOTR\\
SE-SSD & Det3D \cite{zhu2019class} & github.com/Vegeta2020/SE-SSD \\ 
Centerpoint & Openpcdet & github.com/tianweiy/CenterPoint-KITTI \\ 
\bottomrule
\end{tabular}
}
\end{table}

% \subsection{Experimental Set-ups}
% \label{sec: Experimental Set-ups}

% In the Empirical Study, all detectors are executed based on the open-source codes released on GitHub, as shown in Table \ref{tab: det_open}. The configuration files and pre-trained checkpoints can be found in the URLs in Table \ref{tab: det_open}.

% \begin{table*}[]
% \centering
% \caption{{Code sources of detectors}}
% \label{tab: det_open}
% % \resizebox{0.8\textwidth}{!}{%
% \scriptsize
% \begin{tabular}{c|c|l}
% \toprule
% \textbf{Detector} & \textbf{Platform} & \multicolumn{1}{c}{\textbf{URL}} \\ \midrule
% \textbf{SECOND} & Openpcdet\cite{openpcdet2020}  & https://github.com/open-mmlab/OpenPCDet \\
% \textbf{PointRCNN} & Openpcdet  & https://github.com/open-mmlab/OpenPCDet \\
% \textbf{PVRCNN} & Openpcdet & https://github.com/open-mmlab/OpenPCDet \\
% \textbf{BtcDet} & Openpcdet &  https://github.com/xharlie/btcdet\\
% \textbf{VoTr-SSD} & Openpcdet &  https://github.com/PointsCoder/VOTR \\
% \textbf{VoTr-TSD} & Openpcdet &  https://github.com/PointsCoder/VOTR\\
% \textbf{SE-SSD} & Det3D\cite{zhu2019class} & https://github.com/Vegeta2020/SE-SSD \\ 
% \textbf{Centerpoint} & Openpcdet & https://github.com/tianweiy/CenterPoint-KITTI \\ 
% \bottomrule
% \end{tabular}
% \end{table*}

\begin{table*}[hb]
\centering
\caption{$CR_{FC}(\%)$ of detectors on \textit{Car} detection under common corruptions}
\label{tab: FC_BR}
\resizebox{0.9\textwidth}{!}{%
\begin{tabular}{c|c|c|cccccccc|c}
\toprule
\multicolumn{3}{c}{\textbf{Corruption}} & \textbf{PVRCNN} & \textbf{PointRCNN} & \textbf{SECOND} & \textbf{BtcDet} & \textbf{VoTr-SSD} & \textbf{VoTr-TSD} & \textbf{SE-SSD} & \textbf{Centerpoint} & \textbf{Average} \\ \midrule
\multirow{13}{*}{\textbf{Scene-level}} & \multirow{3}{*}{\textbf{Weather}} & \textit{rain} & 1.90 & 1.27 & 0.61 & -0.01 & -0.09 & 0.06 & -0.01 & 2.29 & 0.75 \\
 &  & \textit{snow} & 2.17 & 2.01 & 0.71 & 0.03 & 0.17 & 0.12 & -0.01 & 2.59 & 0.97 \\
 &  & \textit{fog} & 0.03 & 0.02 & 0.10 & 0.02 & 0.21 & 0.16 & 0.04 & 0.16 & 0.09 \\ \cmidrule(r){2-12} 
 & \multirow{5}{*}{\textbf{Noise}} & \textit{uniform\_rad} & 1.35 & 0.82 & 1.27 & 0.00 & -0.06 & 0.00 & -0.01 & 1.29 & 0.58 \\
 &  & \textit{gaussian\_rad} & 1.83 & 1.25 & 1.67 & 0.00 & -0.06 & 0.01 & 0.00 & 1.62 & 0.79 \\
 &  & \textit{impulse\_rad} & 0.12 & 0.01 & 0.14 & 0.02 & 0.07 & 0.06 & 0.04 & 0.25 & 0.09 \\
 &  & \textit{background} & -0.18 & 0.20 & -0.19 & 0.01 & -0.22 & -0.01 & 0.04 & -0.20 & -0.07 \\
 &  & \textit{upsample} & -0.01 & 0.03 & 0.03 & 0.01 & -0.15 & -0.02 & 0.01 & -0.01 & -0.01  \\ \cmidrule(r){2-12} 
 & \multirow{5}{*}{\textbf{Density}} & \textit{cutout} & 0.21 & 0.13 & 0.16 & 0.03 & 0.18 & 0.15 & 0.04 & 0.12 & 0.13 \\
 &  & \textit{local\_dec} & 0.70 & -0.40 & 0.47 & 0.05 & 0.37 & 0.29 & 0.03 & 0.70 & 0.28 \\
 &  & \textit{local\_inc} & 0.04 & 0.02 & 0.05 & 0.01 & 0.00 & 0.05 & 0.05 & 0.04 & 0.03 \\
 &  & \textit{beam\_del} & 0.08 & 0.04 & 0.04 & 0.01 & 0.08 & 0.05 & 0.01 & 0.05 & 0.04 \\
 &  & \textit{layer\_del} & 0.13 & 0.07 & 0.09 & 0.03 & 0.17 & 0.17 & 0.04 & 0.09 & 0.10 \\ \midrule
\multirow{12}{*}{\textbf{Object-level}} & \multirow{4}{*}{\textbf{Noise}} & \textit{uniform} & 0.53 & 0.24 & 0.39 & 0.01 & -0.01 & 0.00 & 0.00 & 0.32 & 0.19 \\
 &  & \textit{gaussian} & 0.71 & 0.32 & 0.50 & 0.01 & -0.01 & 0.00 & 0.01 & 0.41 & 0.24 \\
 &  & \textit{impulse} & 0.03 & 0.00 & 0.04 & -0.01 & 0.00 & 0.02 & 0.00 & 0.00 & 0.01 \\
 &  & \textit{upsample} & 0.06 & 0.00 & 0.12 & 0.00 & 0.01 & 0.02 & 0.03 & 0.04 & 0.03 \\ \cmidrule(r){2-12} 
 & \multirow{3}{*}{\textbf{Density}} & \textit{cutout} & 0.38 & 0.03 & 0.51 & 0.08 & 0.49 & 0.41 & 0.04 & 0.40 & 0.29 \\
 &  & \textit{local\_dec} & 0.26 & -0.03 & 0.40 & 0.07 & 0.37 & 0.31 & 0.04 & 0.32 & 0.22 \\
 &  & \textit{local\_inc} & 0.23 & 0.12 & 0.29 & 0.00 & -0.13 & -0.02 & 0.01 & 0.17 & 0.08 \\ \cmidrule(r){2-12} 
 & \multirow{5}{*}{\textbf{Transformation}} & \textit{shear} & 0.16 & -0.01 & 0.25 & 0.01 & 0.17 & 0.06 & 0.01 & 0.16 & 0.10 \\
 &  & \textit{FFD} & 0.08 & 0.02 & 0.18 & 0.01 & 0.08 & 0.06 & 0.03 & 0.11 & 0.07 \\
 &  & \textit{rotation} & -0.01 & 0.02 & 0.01 & 0.01 & 0.00 & 0.01 & 0.01 & 0.00 & 0.01 \\
 &  & \textit{scaling} & 0.02 & 0.00 & 0.03 & 0.01 & 0.02 & 0.01 & 0.03 & 0.00 & 0.01 \\
 &  & \textit{translation} & 0.14 & 0.09 & 0.14 & 0.03 & 0.10 & 0.09 & 0.05 & 0.15 & 0.10 \\ \midrule
\multicolumn{3}{c|}{mCR} & 0.44 & 0.25 & 0.32 & 0.02 & 0.07 & 0.08 & 0.02 & 0.44 & 0.20 \\ \bottomrule
\end{tabular}%
}
\end{table*}

\subsection{Effects of Common Corruptions to Point Cloud Detectors}
\label{sec: Effects of Common Corruptions to Point Cloud Detectors}

As shown in Table S1 in the \textbf{supplementary}, the average $mCE_{AP}$ of $8.18\%$ still anticipates a noticeable accuracy drop of \textit{Pedestrian} detectors against diverse corruption patterns.
% sever and light corruptions
Specifically, scene-level \{\textit{uniform\_rad, gaussian\_rad, local\_dec}\} and object-level \{\textit{cutout}\} corruptions have the AP loss of more than $20\%$, which presents a serious degradation of detection accuracy. By contrast, some corruption patterns (\eg scene-level \{\textit{background, beam\_del}\}, object-level \{\textit{upsample, rotation, translation}\}) show less effects on detectors (absolute value of $CE_{AP}$ less than $1.25\%$), demonstrating that background and locally upsampling noise, sparse beam loss, and slight rotation and translation don't affect \textit{Pedestrian} detectors' accuracy. Surprisingly, compared to \textit{Car} detection, \textit{Pedestrian} detection are much less affected by \textit{rain} and \textit{snow}, presented by the average $CE_{AP}$ of $2.42\%$ and $2.58\%$. After the investigation of point clouds, we found it is because the proportion ($58.94\%$) of points with zero-value reflection intensity on \textit{Car} objects is much higher than that ($10.01\%$) of pedestrians, and those points with zero-value reflection intensity are easy to be blocked by dense rain and snow droplets.

% Table \ref{tab: CE_Recalls of different detectors and CCs} shows the recall loss (\ie, $CE_{recall}$) of different detectors under different corruptions on \textit{Car} detection. As shown in Table \ref{tab: CE_Recalls of different detectors and CCs}, there is a severe decrease of recall of over $20\%$ under scene-level \{\textit{rain, snow}\} and object-level \{\textit{shear, translation}\} and a slight decrease of recall of less than $1\%$ under scene-level \textit{beam\_del} and object-level \textit{upsample} and a recall decrease of over $9\%$ falls on all detectors.

Table S2 in the \textbf{supplementary} shows the $CE_{AP}$ under different severity levels of corruptions on \textit{Pedestrian} detection. According to Table S2, albeit with some minor exceptions, $CE_{AP}$ of each corruption increases as the severity level increases, which especially rigorously applies to those relatively severe corruptions with the average $CE_{AP}$ of more than $5\%$.

\subsection{Reacts of Detector Designing to Common Corruptions}
\label{sec: Reacts of Detector Designing to Common Corruptions}

% Figure \ref{fig: Average_CE_cor_ProArc} depicts the average precision (AP) loss (\ie, $CE_{AP}$) of one-stage and two-stage detectors on \textit{Car} detection under different common corruptions. As shown in Figure \ref{fig: Average_CE_cor_ProArc}, two-stage detectors present more accurate detection under the scene-level \{\textit{cutout, local\_dec, beam\_del, layer\_del}\} and \{\textit{shear, scale}\}, displaying a lower average $CE_{AP}$, while one-stage detectors present more accurate under the rest of common corruptions. The details of $CE_{AP}$ of
% different detectors under different corruptions are recorded in Table \ref{tab: CEs of different proposal architectures}. 
Figure S1 in the \textbf{supplementary} depicts the relationship between of $AP$ and $mCE_{AP}$ of \textit{Pedestrian} detectors.
As shown in Figure S1, similar to \textit{Car} detection, the $mCE_{AP}$ of \textit{Pedestrian} detection increases as its $AP$ increases.

\subsection{Detection Bugs in Detectors under Common Corruptions}
\label{sec: Detection Bugs in Detectors under Common Corruptions}

More details of the increase of bug rates (\ie, $CR$) of different detectors under different corruptions are recorded in Table \ref{tab: FC_BR}, Table \ref{tab: FD_BR}, and Table \ref{tab: MD_BR}.

\begin{table*}[]
\centering
\caption{$CR_{FD}(\%)$ of detectors on \textit{Car} detection under common corruptions}
\label{tab: FD_BR}
\resizebox{0.9\textwidth}{!}{%
\begin{tabular}{c|c|c|cccccccc|c}
\toprule
\multicolumn{3}{c}{\textbf{Corruption}} & \textbf{PVRCNN} & \textbf{PointRCNN} & \textbf{SECOND} & \textbf{BtcDet} & \textbf{VoTr-SSD} & \textbf{VoTr-TSD} & \textbf{SE-SSD} & \textbf{Centerpoint} & \textbf{Average} \\ \midrule
\multirow{13}{*}{\textbf{Scene-level}} & \multirow{3}{*}{\textbf{Weather}} & \textit{rain} & 8.25 & 4.23 & 5.70 & 7.90 & 13.39 & 11.46 & 4.78 & 5.24 & 7.62 \\
 &  & \textit{snow} & 8.54 & 7.45 & 5.93 & 14.51 & 12.90 & 16.34 & 10.11 & 5.39 & 10.15 \\
 &  & \textit{fog} & 0.82 & 0.11 & 0.62 & -0.30 & -0.84 & -0.49 & 0.10 & 0.71 & 0.09 \\ \cmidrule(r){2-12} 
 & \multirow{5}{*}{\textbf{Noise}} & \textit{uniform\_rad} & 4.08 & 5.94 & 3.30 & 3.81 & 2.61 & 4.01 & 5.52 & 2.50 & 3.97 \\
 &  & \textit{gaussian\_rad} & 4.88 & 6.83 & 3.87 & 4.31 & 3.34 & 4.94 & 6.44 & 3.02 & 4.70 \\
 &  & \textit{impulse\_rad} & 1.54 & 2.30 & 0.99 & 1.81 & 2.55 & 3.72 & 1.22 & 0.92 & 1.88 \\
 &  & \textit{background} & -0.34 & 2.25 & -2.25 & 0.34 & -3.22 & -0.14 & 0.78 & -2.14 & -0.59 \\
 &  & \textit{upsample} & -0.04 & 0.82 & -0.34 & 0.34 & 0.29 & 0.14 & 0.69 & -0.44 & 0.18 \\ \cmidrule(r){2-12} 
 & \multirow{5}{*}{\textbf{Density}} & \textit{cutout} & 0.73 & 0.57 & 0.58 & 1.11 & 1.05 & 1.17 & 1.87 & 0.41 & 0.94 \\
 &  & \textit{local\_dec} & 3.79 & - & 2.77 & 3.40 & 3.44 & 4.29 & 4.43 & 2.83 & 3.56 \\
 &  & \textit{local\_inc} & 0.39 & 0.95 & 0.23 & 0.43 & 0.53 & 0.55 & 0.48 & 0.30 & 0.48 \\
 &  & \textit{beam\_del} & 0.39 & 0.06 & 0.39 & 0.54 & 0.52 & 0.48 & 0.75 & 0.30 & 0.43 \\
 &  & \textit{layer\_del} & 0.98 & 0.54 & 0.79 & 1.25 & 0.98 & 1.01 & 1.36 & 0.73 & 0.95 \\ \midrule
\multirow{12}{*}{\textbf{Object-level}} & \multirow{4}{*}{\textbf{Noise}} & \textit{uniform} & 2.30 & 4.51 & 1.25 & 2.85 & 1.37 & 2.13 & 3.01 & 1.10 & 2.31 \\
 &  & \textit{gaussian} & 3.11 & 6.00 & 1.70 & 3.99 & 1.98 & 3.11 & 3.91 & 1.47 & 3.16 \\
 &  & \textit{impulse} & 0.89 & 1.47 & 0.53 & 1.40 & 1.06 & 1.32 & 1.43 & 0.53 & 1.08 \\
 &  & \textit{upsample} & 0.28 & 0.79 & 0.22 & 0.33 & 0.24 & 0.13 & 0.95 & 0.16 & 0.39 \\ \cmidrule(r){2-12} 
 & \multirow{3}{*}{\textbf{Density}} & \textit{cutout} & 2.29 & -0.92 & 1.64 & 0.39 & 2.75 & 3.27 & 0.54 & 0.77 & 1.34 \\
 &  & \textit{local\_dec} & 1.83 & -1.15 & 1.27 & 0.10 & 2.02 & 2.57 & -0.13 & 0.48 & 0.87 \\
 &  & \textit{local\_inc} & 3.54 & 5.44 & 2.54 & 4.81 & 3.43 & 4.74 & 6.77 & 2.31 & 4.20 \\ \cmidrule(r){2-12} 
 & \multirow{5}{*}{\textbf{Transformation}} & \textit{shear} & 10.08 & 17.48 & 7.56 & 23.27 & 11.70 & 14.84 & 24.29 & 7.74 & 14.62 \\
 &  & \textit{FFD} & 8.06 & 14.85 & 5.59 & 18.56 & 8.82 & 12.28 & 20.52 & 5.95 & 11.83 \\
 &  & \textit{rotation} & 0.22 & 0.13 & 0.12 & 0.04 & 0.25 & 0.29 & 0.41 & 0.07 & 0.24 \\
 &  & \textit{scaling} & 2.31 & 3.82 & 1.81 & 4.40 & 3.24 & 3.38 & 5.37 & 1.90 & 3.28 \\
 &  & \textit{translation} & 1.58 & 1.09 & 0.87 & 1.55 & 2.78 & 3.24 & 1.09 & 0.88 & 1.63 \\ \midrule
\multicolumn{3}{c|}{mCR} & 2.82 & 3.56 & 1.91 & 4.06 & 3.09 & 3.95 & 4.27 & 1.73 & 3.17 \\ \bottomrule
\end{tabular}%
}
\end{table*}

\begin{table*}[hb]
\centering
\caption{$CR_{MD}(\%)$ of detectors on \textit{Car} detection under common corruptions}
\label{tab: MD_BR}
\resizebox{0.9\textwidth}{!}{%
\begin{tabular}{c|c|c|cccccccc|c}
\toprule
\multicolumn{3}{c}{\textbf{Corruption}} & \textbf{PVRCNN} & \textbf{PointRCNN} & \textbf{SECOND} & \textbf{BtcDet} & \textbf{VoTr-SSD} & \textbf{VoTr-TSD} & \textbf{SE-SSD} & \textbf{Centerpoint} & \textbf{Average} \\ \midrule
\multirow{13}{*}{\textbf{Scene-level}} & \multirow{3}{*}{\textbf{Weather}} & \textit{rain} & -9.33 & -4.63 & -3.42 & -12.61 & -8.37 & -5.64 & -11.01 & -4.89 & -7.49 \\
 &  & \textit{snow} & -0.71 & -3.47 & 0.62 & -8.96 & 4.14 & -0.61 & -9.99 & -0.40 & -2.42 \\
 &  & \textit{fog} & -0.60 & -0.43 & 0.25 & -0.54 & 6.77 & 6.53 & 2.70 & -1.59 & 1.64 \\ \cmidrule(r){2-12} 
 & \multirow{5}{*}{\textbf{Noise}} & \textit{uniform\_rad} & -2.10 & -5.78 & -2.98 & -4.51 & -2.79 & -3.62 & -4.63 & -2.45 & -3.61 \\
 &  & \textit{gaussian\_rad} & -2.32 & -6.42 & -3.03 & -5.00 & -3.37 & -4.28 & -5.17 & -2.68 & -4.03 \\
 &  & \textit{impulse\_rad} & 0.26 & -5.51 & 0.85 & -2.23 & -2.45 & -2.89 & -1.51 & 1.24 & -1.53 \\
 &  & \textit{background} & 7.43 & -7.01 & 13.13 & -1.12 & 19.16 & 10.67 & 5.10 & 12.48 & 7.48 \\
 &  & \textit{upsample} & 2.07 & -0.73 & 2.88 & -0.62 & 0.79 & 1.20 & 1.07 & 2.80 & 1.18 \\ \cmidrule(r){2-12} 
 & \multirow{5}{*}{\textbf{Density}} & \textit{cutout} & 1.74 & 1.83 & 1.59 & -1.01 & 1.77 & 0.53 & -1.03 & 2.26 & 0.96 \\
 &  & \textit{local\_dec} & -0.08 & - & -0.09 & -3.24 & 2.18 & -0.38 & -3.44 & 0.34 & -0.67 \\
 &  & \textit{local\_inc} & 0.39 & -1.11 & 0.59 & -0.37 & 1.05 & 1.16 & 0.81 & 0.29 & 0.35 \\
 &  & \textit{beam\_del} & -0.37 & 0.14 & -0.48 & -1.02 & -0.37 & -0.92 & -1.27 & -0.22 & -0.56 \\
 &  & \textit{layer\_del} & 0.13 & 1.32 & -0.27 & -1.00 & 1.02 & 0.55 & -0.99 & 0.20 & 0.12 \\ \midrule
\multirow{12}{*}{\textbf{Object-level}} & \multirow{4}{*}{\textbf{Noise}} & \textit{uniform} & 0.90 & -0.93 & 0.32 & 0.86 & -0.15 & -0.31 & 0.06 & 0.50 & 0.16 \\
 &  & \textit{gaussian} & 1.23 & -1.00 & 0.49 & 1.33 & -0.11 & -0.32 & 0.27 & 0.69 & 0.32 \\
 &  & \textit{impulse} & 0.13 & -0.74 & 0.10 & 0.06 & 0.01 & -0.16 & 0.15 & 0.22 & -0.03 \\
 &  & \textit{upsample} & -0.02 & -1.15 & 0.01 & -0.22 & -0.07 & -0.11 & -0.09 & 0.05 & -0.20 \\ \cmidrule(r){2-12} 
 & \multirow{3}{*}{\textbf{Density}} & \textit{cutout} & 5.05 & 10.70 & 3.62 & 5.00 & 3.76 & 4.81 & 5.15 & 4.74 & 5.35 \\
 &  & \textit{local\_dec} & 4.76 & 9.90 & 3.36 & 4.70 & 3.57 & 4.52 & 4.89 & 4.43 & 5.02 \\
 &  & \textit{local\_inc} & 0.52 & 0.05 & 0.30 & 0.86 & 0.24 & 0.23 & 0.85 & 0.45 & 0.44 \\ \cmidrule(r){2-12} 
 & \multirow{5}{*}{\textbf{Transformation}} & \textit{shear} & 0.01 & -0.60 & 0.05 & 0.40 & -0.09 & -0.18 & 0.05 & 0.22 & -0.02 \\
 &  & \textit{FFD} & 0.08 & -1.11 & 0.04 & 0.72 & 0.02 & 0.05 & 1.08 & 0.24 & 0.14 \\
 &  & \textit{rotation} & 0.03 & -0.28 & 0.00 & 0.01 & -0.06 & -0.05 & -0.02 & 0.03 & -0.04 \\
 &  & \textit{scaling} & 0.08 & 0.08 & 0.01 & 0.10 & -0.03 & -0.04 & 0.31 & 0.09 & 0.07 \\
 &  & \textit{translation} & 0.29 & -1.29 & 0.45 & 0.27 & -0.07 & -0.23 & -0.02 & 0.60 & 0.00\\ \midrule
\multicolumn{3}{c|}{mCR} & 0.38 & -0.76 & 0.74 & -1.13 & 1.06 & 0.42 & -0.67 & 0.79 & 0.11 \\ \bottomrule
\end{tabular}%
}
\end{table*}

According to Figure \ref{fig: TD_dis}, the TD rate of \textit{Car} detection under clean and corrupted (at the severity level of 3) observations \wrt the distance range of [0, 30]m remains approximately still. However, beyond this distance range (\ie, distance > 30m), the TD rate decreases as the distance increases. On the other hand, compared to the clean observation, the TD rates under corrupted observations are always lower at any distance. More details of each corruption are shown in Figure S2 and S3 in the \textbf{supplementary}.

% \begin{figure*}[h]
%   \centering
%   \includegraphics[width=0.8\linewidth]{image/TD_distance_scene.png}
%   \caption{TD rate of all frames under scene-level corruptions \wrt different distances of objects to the LiDAR sensor ({green} dotted lines for the median and {green} triangles for the mean)}
%   \label{fig: TD_dis_scene}
% \end{figure*}

% \begin{figure*}[h]
%   \centering
%   \includegraphics[width=0.8\linewidth]{image/TD_distance_object.png}
%   \caption{TD rate of all frames under object-level corruptions \wrt different distances of objects to the LiDAR sensor ({green} dotted lines for the median and {green} triangles for the mean)}
%   \label{fig: TD_dis_object}
% \end{figure*}

\subsection{Robustness Enhancement by Data Augmentation and Denoising}
\label{sec: Robustness Enhancement by Data Augmentation and Denoising}

More details of $CE_{AP}$ of different detectors on \textit{Car} detection with PA-DA (data augmentation) and/or KNN-RO (denoising) under different corruptions are shown in Table S3, Table S4, and Table S5 in the \textbf{supplementary}.

\begin{table}[]
\centering
\caption{Average $CE_{AP}(\%)$ of detectors given different common corruptions on \textit{Pedestrian} detection with DA and/or denoising}
\label{tab: CE given different common corruptions with DA and/or Denoising on pedestrian detection}
\resizebox{0.5\textwidth}{!}{%
\begin{tabular}{c|c|c|cccc}
\toprule
\multicolumn{3}{c|}{\textbf{Corruption}} & \textbf{Origin} & \textbf{PA-DA} & \textbf{KNN-RO} & \begin{tabular}{c}\textbf{PA-DA} +\\\textbf{KNN-RO} \end{tabular} \\ \midrule
\multirow{13}{*}{\rotatebox{90}{\textbf{Scene-level}}} & \multirow{3}{*}{\textbf{Weather}} & \textit{rain} & 2.42 & 1.67(-0.75) & 1.6(-0.82) & 1.39(-1.03) \\
 &  & \textit{snow} & 2.58 & 3.99(+1.41) & 2.0(-0.58) & 3.24(+0.66) \\
 &  & \textit{fog} & 6.27 & 4.82(-1.45) & 6.17(-0.1) & 4.22(-2.05) \\ \cmidrule(r){2-7} 
 & \multirow{5}{*}{\textbf{Noise}} & \textit{uniform\_rad} & 30.14 & 25.13(-5.01) & 29.79(-0.35) & 24.93(-5.21) \\
 &  & \textit{gaussian\_rad} & 36.99 & 32.22(-4.77) & 36.8(-0.19) & 32.32(-4.67) \\
 &  & \textit{impulse\_rad} & 2.09 & 5.1(+3.01) & 2.21(+0.12) & 5.1(+3.01) \\
 &  & \textit{background} & 0.62 & 0.09(-0.53) & -0.22(-0.84) & -0.29(-0.91) \\
 &  & \textit{upsample} & 1.38 & 1.77(+0.39) & 1.24(-0.14) & 1.71(+0.33) \\ \cmidrule(r){2-7} 
  & \multirow{5}{*}{\textbf{Density}} & \textit{cutout} & 7.43 & 6.02(-1.41) & 7.3(-0.13) & 5.73(-1.7) \\
 &  & \textit{local\_dec} & 12.6 & 12.25(-0.35) & 12.62(+0.02) & 12.21(-0.39) \\
 &  & \textit{local\_inc} & 1.36 & 1.42(+0.06) & 1.49(+0.13) & 1.23(-0.13) \\
 &  & \textit{beam\_del} & -0.02 & 0.23(+0.25) & -0.15(-0.13) & -0.13(-0.11) \\
 &  & \textit{layer\_del} & 3.81 & 2.97(-0.84) & 3.64(-0.17) & 2.66(-1.15) \\ \midrule
\multirow{12}{*}{\rotatebox{90}{\textbf{Object-level}}} & \multirow{4}{*}{\textbf{Noise}} & \textit{uniform} & 4.21 & 3.07(-1.14) & 4.25(+0.04) & 2.98(-1.23) \\
 &  & \textit{gaussian} & 5.47 & 4.09(-1.38) & 5.59(+0.12) & 4.04(-1.43) \\
 &  & \textit{impulse} & 1.85 & 1.74(-0.11) & 2.05(+0.2) & 1.79(-0.06) \\
 &  & \textit{upsample} & 0.08 & -0.21(-0.29) & -0.14(-0.22) & -0.37(-0.45) \\ \cmidrule(r){2-7} 
 & \multirow{3}{*}{\textbf{Density}} & \textit{cutout} & 20.34 & 18.15(-2.19) & 20.07(-0.27) & 18.04(-2.3) \\
 &  & \textit{local\_dec} & 17.05 & 15.71(-1.34) & 17.09(+0.04) & 15.71(-1.34) \\
 &  & \textit{local\_inc} & 8.21 & 5.09(-3.12) & 8.03(-0.18) & 5.09(-3.12) \\ \cmidrule(r){2-7} 
 & \multirow{5}{*}{\textbf{\tiny Transformation}} & \textit{shear} & 16.91 & 13.22(-3.69) & 16.85(-0.06) & 12.87(-4.04) \\
 &  & \textit{FFD} & 11.74 & 9.5(-2.24) & 11.7(-0.04) & 9.34(-2.4) \\
 &  & \textit{rotation} & -0.03 & 0.23(+0.26) & -0.09(-0.06) & 0.05(+0.08) \\
 &  & \textit{scaling} & 3.73 & 3.98(+0.25) & 3.65(-0.08) & 3.67(-0.06) \\
 &  & \textit{translation} & -1.22 & -1.42(-0.2) & -1.08(+0.14) & -1.34(-0.12) \\ \midrule
\multicolumn{3}{c|}{Average} & 7.84 & 6.83(-1.01) & 7.7(-0.14) & 6.65(-1.19) \\ \bottomrule
\end{tabular}
}
\end{table}

Table \ref{tab: CE given different common corruptions with DA and/or Denoising on pedestrian detection} gives the details of average AP loss (\ie, $CE_{AP}$) given different corruptions on \textit{Pedestrian} detection with PA-DA (data augmentation) and/or KNN-RO (denoising). According to Table \ref{tab: CE given different common corruptions with DA and/or Denoising on pedestrian detection}, PA-DA presents an AP increase of over $3\%$ on \textit{Pedestrian} detection under scene-level \{\textit{uniform\_rad, gaussian\_rad}\} and object-level \{\textit{local\_inc, shear}\}, and PA-DA + KNN-RO presents an AP increase of over $3\%$ under scene-level \{\textit{uniform\_rad, gaussian\_rad}\} and object-level \{\textit{local\_inc, shear}\}.

\begin{table}[]
\centering
\caption{Average $CE_{AP}(\%)$ of different detectors on \textit{Pedestrian} detection with DA and/or denoising}
\label{tab: CE given different detectors with DA and/or Denoising on pedestrian detection}
\resizebox{0.5\textwidth}{!}{%
\begin{tabular}{c|cccc}
\toprule
\textbf{Detector} & \textbf{Origin} & \textbf{PA-DA} & \textbf{KNN-RO} & \textbf{PA-DA + KNN-RO} \\ \midrule
\textbf{PVRCNN} & 8.71 & 10.6(+1.89) & 8.2(-0.51) & 10.05(+1.34) \\
\textbf{PointRCNN} & 7.08 & 3.73(-3.35) & 7.05(-0.03) & 3.57(-3.51) \\
\textbf{SECOND} & 8.38 & 7.63(-0.75) & 8.46(+0.08) & 7.76(-0.62) \\ 
\textbf{Centerpoint} & 7.19 & 5.37(-1.82) & 7.91(+0.72) & 5.21(-1.98) \\ 
\midrule
Average & 7.84 & 6.83(-1.01) & 7.7(-0.14) & 6.65(-1.19) \\
\bottomrule
\end{tabular}%
}
\end{table}

Table \ref{tab: CE given different detectors with DA and/or Denoising on pedestrian detection} shows $CE_{AP}$ of different detectors on \textit{Pedestrian} detection with PA-DA (data augmentation) and/or KNN-RO (denoising). According to Table \ref{tab: CE given different detectors with DA and/or Denoising on pedestrian detection}, with only PA-DA, an increase of average precision (AP) falls on PointRCNN, SECOND, and Centerpoint; with only KNN-RO, an increase of AP falls on PVRCNN and PointRCNN; with PA-DA + KNN-RO, an AP increase falls on PointRCNN, SECOND, and Centerpoint.

% \clearpage

\section{Corruption Simulation Naturalness Validation}

\subsection{Naturalness Validation of Weather Corruption Simulation}
\label{sec: naturalness validation for weather}

\subsubsection{Snow and Fog Validation}
% summary
% step
{
To verify the naturalness of snow and fog simulation, we train weather-oriented PointNet-based classifiers via data collected in real snowy and foggy weather (from public datasets as in Table \ref{tab: weather classification}). 
}

{
{\noindent \bf Dataset Split:} As for snow or fog, we randomly divide collected real corrupted and clean data into training set $D_{train}$ and validation set $D_{val}$ at the size ratio of 9:1, and gather the original clean and simulated corrupted KITTI data as testing set $D_{test}$. Note that, $D_{train}$, $D_{val}$, and $D_{test}$ all contain roughly $50\%$ corrupted data and $50\%$ clean data. 
}

{
{\noindent \bf Training and Testing:} We first train the PointNet model with $D_{train}$ to learn the model to classify the clean or corrupted condition of point clouds. By means of $D_{val}$, we validate the model across all epochs and settle its best checkpoint. The settled model is utilized to classify the clean or corrupted condition of point clouds in $D_{test}$. 
}

{
% PointNet-related experiment setting
{\noindent \bf Experiment Setting:} As for the experiment setting, we set the epoch number to 80 and the batch size to 32 for the model training and validation. The training and testing are all executed on the NVIDIA RTX A6000 GPU with a memory of 48GB. We choose PointNet as the classification model due to its efficiency and effectiveness in recognizing the global point feature in point clouds \cite{qi2017pointnet}, which fits into the global effects of weather corruptions on LiDAR scanning. 
% For snow validation, the maximum range of 15m and the maximum point number of 12000 are used to pre-process and normalize input data before training and testing. For fog validation, corresponding parameters are reduced to 10m and 5000 for model regularization to ease the "overfitting" problem caused by relatively small data size \ref{tab: weather classification}. 
}

% validation result and illumination
{
{\noindent \bf Analysis:} According to the snow classification in \ref{tab: weather classification}, the validation accuracy of $99.11\%$ illustrates the snow classifier's precise snow recognition on point clouds collected in the real world. Hence, the classifier's testing accuracy of $97.13\%$ on simulated data demonstrates the high naturalness of the snow simulation. Likewise, even though the fog classifier's validation accuracy is reduced by the small data size, the testing accuracy of $92.60\%$ on simulated data still presents a high similarity of data corrupted by simulated fog to data affected by real fog.}

\subsubsection{Rain Validation}
% rain validation
In Figure \ref{fig: rain_comparison}, data of real clean and real rain from Boreas were sampled at the same location but at different time-stamps; the data of simulated rain was augmented on the basis of the data of real clean. 
According to Figure \ref{fig: rain_comparison}, compared with the data of real clean, the point cloud under simulated rain has sparse ``false points'' (as in red boxes) surrounding the LiDAR sensor nearby and wipes out some points on the road. Both are similar to the effects of real-world rain, shown by the yellow boxes and missing points on the road. {More comparisons between real rainy data and simulated rainy data are shown in Figure S4 in the \textbf{supplementary}.}

% \begin{figure}[h]
%   \centering
%   \includegraphics[width=1\linewidth]{image/rain_comparison.png}
%   \caption{Comparison between real \textit{rain} and simulated \textit{rain} ({\color{red}red} and {\color{yellow}yellow} boxes contain the false points in the simulated and real \textit{rain}, respectively; the data of real \textit{clean} and real \textit{rain} from Boreas were sampled at the same location; the simulated \textit{rain} data was augmented on the basis of the real \textit{clean} data)}
%   \label{fig: rain_comparison}
% \end{figure}

% \begin{figure*}[h]
%   \centering
%   \includegraphics[width=0.7\linewidth]{image/rain_comparison2.png}
%   \caption{{More visualizations of real \textit{rain} and simulated \textit{rain}}}
%   \label{fig: rain_comparison2}
% \end{figure*}

\section{Implementation of Corruption Simulation}
\label{sec: corruption_simulation_implementation}

Figures S5 to S30 in the \textbf{supplementary} display the point cloud examples under ``clean'' and simulated common corruptions (at severity level of 3). Those examples are based on the KITTI LiDAR example with ID = ``000008''. Besides, we provide the ground-truth annotations of objects and detection results obtained by PVRCNN. Taking Figures S6 to S30 for examples, we next detail on the simulation implementation. 

% %% corruption display
% \begin{figure}[h]
%   \centering
%   \includegraphics[width=1\linewidth]{image/corruptions/clean.png}
%   \caption{\textit{Clean} data and the \textit{Car} detection by PVRCNN ({\color{red}Red} BBoxes for the ground-truth and {\color{green}green} ones for the PVRCNN detection)}
%   \label{fig: corruptions_clean}
% \end{figure}

\subsection{Scene-level Corruption}
{
 {\noindent \bf \textit{Rain} and \textit{Snow}.} We adopt the rain and snow simulators of LISA \cite{kilic2021lidar} to simulate rain and snow corruption. For LISA, the parameters \textit{rainfall rate} and \textit{snow rate} can be regulated to simulate corruptions at different severity levels. After the investigation of the real-world rainfall rates \cite{enwiki:1102188829}, we set \textit{rainfall rate} and \textit{snow rate} to $\{0, 5.0, 15.0, 50.0, 150.0, 500.0\}$ mm/hr and $\{0, 0.5, 1.5, 5.0, 15.0, 50.0\}$ mm/hr as 6 severity levels (\ie, 0 to 5) for rain and snow corruption, respectively. Figures S6 and S7 display point cloud examples under \textit{rain} and \textit{snow}. \textit{Note that, the severity level 0 stands for the original clean data, which also applies to the rest of the corruptions.}
 }

% \begin{figure}[h]
%   \centering
%   \includegraphics[width=1\linewidth]{image/corruptions/rain.png}
%   \caption{ Point cloud under \textit{rain} corruption and the \textit{Car} detection by PVRCNN ({\color{red}Red} BBoxes for the ground-truth and {\color{green}green} ones for the PVRCNN detection)}
%   \label{fig: corruptions_rain}
% \end{figure}

% \begin{figure}[h]
%   \centering
%   \includegraphics[width=1\linewidth]{image/corruptions/snow.png}
%   \caption{ Point cloud under \textit{snow} corruption and the \textit{Car} detection by PVRCNN ({\color{red}Red} BBoxes for the ground-truth and {\color{green}green} ones for the PVRCNN detection)}
%   \label{fig: corruptions_snow}
% \end{figure}

{
{\noindent \bf \textit{Fog}.} We adopt the fog simulator of LFS \cite{hahner2021fog} to simulate fog corruption in point clouds. For LFS, the parameter $\alpha$ is set to regulate the corruption severity levels. Following the recommended setting in \cite{hahner2021fog} (with slight modifications for the relatively wide range of severity), we set $\alpha$ to $\{0, 0.005, 0.01, 0.02, 0.05, 0.1\}$. Figure S8 displays the point cloud example under fog.
}

% \begin{figure}[h]
%   \centering
%   \includegraphics[width=1\linewidth]{image/corruptions/fog.png}
%   \caption{ Point cloud under \textit{fog} corruption and the \textit{Car} detection by PVRCNN ({\color{red}Red} BBoxes for the ground-truth and {\color{green}green} ones for the PVRCNN detection)}
%   \label{fig: corruptions_fog}
% \end{figure}

{
{\noindent \bf \textit{Uniform\_rad} and \textit{Gaussian\_rad}.} To fit into the mechanism of LiDAR scanning, we first convert the coordinates of points at from the Cartesian system into the spherical system (\ie, $[x, y, z]$ to $[r, \theta, \phi]$). Then, we add the uniform or Gaussian noise into $r$ of every point. The upper and lower bounds of the range of \textit{uniform\_rad} are set to +/- \{0, 0.04, 0.08, 0.12, 0.16, 0.2\}m and the standard deviation of \textit{gaussian\_rad} to \{0, 0.04, 0.06, 0.08, 0.10, 0.12\}m. Figures S9 and S10 display the examples under \textit{uniform\_rad} and \textit{gaussian\_rad}. 
}

% \begin{figure}[h]
%   \centering
%   \includegraphics[width=1\linewidth]{image/corruptions/uniform_rad.png}
%   \caption{ Point cloud under \textit{uniform\_rad} corruption and the \textit{Car} detection by PVRCNN ({\color{red}Red} BBoxes for the ground-truth and {\color{green}green} ones for the PVRCNN detection)}
%   \label{fig: corruptions_uniform_rad}
% \end{figure}

% \begin{figure}[h]
%   \centering
%   \includegraphics[width=1\linewidth]{image/corruptions/gaussian_rad.png}
%   \caption{ Point cloud under \textit{gaussian\_rad} corruption and the \textit{Car} detection by PVRCNN ({\color{red}Red} BBoxes for the ground-truth and {\color{green}green} ones for the PVRCNN detection)}
%   \label{fig: corruptions_gaussian_rad}
% \end{figure}

{
{\noindent \bf \textit{Impulse\_rad}.} Likewise, we first convert coordinates of points at from the Cartesian system to the spherical system, and then, we add deterministic perturbations of $+/- 0.2m$ to $r$ of a certain portion of points. The portion is set to $\{0, N/30, N/25, N/20, N/15, N/10\}$ for 6 severity levels, where $N$ represents the number of points. Figure S11 displays the point cloud example under \textit{impulse\_rad}. 
}

% \begin{figure}[h]
%   \centering
%   \includegraphics[width=1\linewidth]{image/corruptions/impulse_rad.png}
%   \caption{ Point cloud under \textit{impulse\_rad} corruption and the \textit{Car} detection by PVRCNN ({\color{red}Red} BBoxes for the ground-truth and {\color{green}green} ones for the PVRCNN detection)}
%   \label{fig: corruptions_impulse_rad}
% \end{figure}

{
{\noindent \bf \textit{Background}.} For \textit{background}, within the spatial range of the scene, background points are randomly sampled uniformly and concatenated to the original points. The number of background points is set to $\{0, N/45, N/40, N/35, N/30, N/20\}$ where $N$ represents the number of original points. Figure S12 displays the point cloud example under \textit{background}. 
}

% \begin{figure}[h]
%   \centering
%   \includegraphics[width=1\linewidth]{image/corruptions/background.png}
%   \caption{ Point cloud under \textit{background} corruption and the \textit{Car} detection by PVRCNN ({\color{red}Red} BBoxes for the ground-truth and {\color{green}green} ones for the PVRCNN detection)}
%   \label{fig: corruptions_background}
% \end{figure}

{
{\noindent \bf \textit{Upsample}.} For \textit{upsample}, we spatially upsample points (with the random bias within [-0.1, 0.1]) nearby a certain portion of original points. The portion is set to $\{0, N/10, N/8, N/6, N/4, N/2\}$ for 6 severity levels, where $N$ represents the number of original points. Figure S13 displays the point cloud example under \textit{upsample}.
}

% \begin{figure}[h]
%   \centering
%   \includegraphics[width=1\linewidth]{image/corruptions/upsample.png}
%   \caption{ Point cloud under \textit{upsample} corruption and the \textit{Car} detection by PVRCNN ({\color{red}Red} BBoxes for the ground-truth and {\color{green}green} ones for the PVRCNN detection)}
%   \label{fig: corruptions_upsample}
% \end{figure}

{
{\noindent \bf \textit{Cutout}.} For \textit{cutout}, we first randomly select a certain portion of points as centers. By KNN, we erase the distance-related neighborhood of every centers. The portion of selected points and the number of neighbor points are set to \{(0,0), (N/2000,100), (N/1500,100), (N/1000,100), (N/800,100), (N/600,100)\}. Figure S14 displays the point cloud example under \textit{cutout}.
}

% \begin{figure}[h]
%   \centering
%   \includegraphics[width=1\linewidth]{image/corruptions/cutout.png}
%   \caption{ Point cloud under \textit{cutout} corruption and the \textit{Car} detection by PVRCNN ({\color{red}Red} BBoxes for the ground-truth and {\color{green}green} ones for the PVRCNN detection)}
%   \label{fig: corruptions_cutout}
% \end{figure}

{
{\noindent \bf \textit{Local\_dec} and \textit{local\_inc}.} For \textit{local\_dec}, we randomly select a certain portion of points as center, and delete the $75\%$ points of the spatial neighborhood of every center. For \textit{local\_inc}, within the neighborhood of every center, we utilize quadratic-polynomial fitting to upsample points as many as the neighbor points and concatenate them into the original points. The portion of selected points and the number of neighbor points are set to \{(0, 0), (N/300,100), (N/250,100), (N/200,100), (N/150,100), (N/100,100)\} for \textit{local\_dec} and \{(0, 0), (N/2000,100), (N/1500,100), (N/1000,100), (N/800,100), (N/600,100)\} for \textit{local\_inc}. Figures S15 and S16 display the point cloud examples under \textit{local\_dec} and \textit{local\_inc}.
}

% \begin{figure}[h]
%   \centering
%   \includegraphics[width=1\linewidth]{image/corruptions/local_dec.png}
%   \caption{ Point cloud under \textit{local\_dec} corruption and the \textit{Car} detection by PVRCNN ({\color{red}Red} BBoxes for the ground-truth and {\color{green}green} ones for the PVRCNN detection)}
%   \label{fig: corruptions_local_dec}
% \end{figure}

% \begin{figure}[h]
%   \centering
%   \includegraphics[width=1\linewidth]{image/corruptions/local_inc.png}
%   \caption{ Point cloud under \textit{local\_inc} corruption and the \textit{Car} detection by PVRCNN ({\color{red}Red} BBoxes for the ground-truth and {\color{green}green} ones for the PVRCNN detection)}
%   \label{fig: corruptions_local_inc}
% \end{figure}

{
{\noindent \bf \textit{Beam\_del}.} For \textit{beam\_del}, we randomly delete a certain portion of points in the point cloud. The portion is set to $\{0, N/100, N/30, N/10, N/5, N/3\}$ for 6 severity levels. Figure S17 displays the point cloud example under \textit{beam\_del}.
}

% \begin{figure}[h]
%   \centering
%   \includegraphics[width=1\linewidth]{image/corruptions/beam_del.png}
%   \caption{ Point cloud under \textit{beam\_del} corruption and the \textit{Car} detection by PVRCNN ({\color{red}Red} BBoxes for the ground-truth and {\color{green}green} ones for the PVRCNN detection)}
%   \label{fig: corruptions_beam_del}
% \end{figure}

{
{\noindent \bf \textit{Layer\_del}.} First, we convert the coordinates of points from the Cartesian system into the spherical system and obtain the range of the polar angle $\theta$ of all points in the point cloud. 
Then, based on the layer number of LiDAR scanning (\eg, 64 for KITTI \cite{geiger2013vision}), we divide the range of $\theta$ into 32 or 64 bins. 
Finally, we randomly select a certain number of $\theta$ bins and delete the corresponding points. 
For KITTI, the number of deleted bins is set to $\{0, 3, 7, 11, 15, 19\}$. Figure S18 displays the point cloud example under \textit{layer\_del}.
}

% \begin{figure}[h]
%   \centering
%   \includegraphics[width=1\linewidth]{image/corruptions/layer_del.png}
%   \caption{ Point cloud under \textit{layer\_del} corruption and the \textit{Car} detection by PVRCNN ({\color{red}Red} BBoxes for the ground-truth and {\color{green}green} ones for the PVRCNN detection)}
%   \label{fig: corruptions_layer_del}
% \end{figure}

\subsection{Object-level Corruption}

{
{\noindent \bf \textit{Uniform\_obj}, \textit{Gaussian\_obj}, and \textit{Impulse\_obj}.} For every annotated object (\eg, Car or Cyclist), we add the noise into the Cartesian coordinates of its points. The upper and lower bounds of the range of \textit{uniform\_obj} are set to $+/- \{0, 0.02, 0.04, 0.06, 0.08, 0.10\}$m. The the standard deviation of \textit{gaussian\_obj}is set to $\{0, 0.02, 0.03, 0.04, 0.05, 0.06\}$m. The number of points affected by \textit{impulse\_obj} with the bias of $+/- 0.1m$ is set to $\{0, N/30, N/25, N/20, N/15, N/10\}$. Figures S19, S20, and S21 display the point cloud examples under \textit{uniform\_obj}, \textit{gaussian\_obj}, and \textit{impulse\_obj}. 
}

% \begin{figure}[h]
%   \centering
%   \includegraphics[width=1\linewidth]{image/corruptions/uniform_obj.png}
%   \caption{ Point cloud under object-level \textit{uniform} corruption and the \textit{Car} detection by PVRCNN ({\color{red}Red} BBoxes for the ground-truth and {\color{green}green} ones for the PVRCNN detection)}
%   \label{fig: corruptions_uniform_obj}
% \end{figure}

% \begin{figure}[h]
%   \centering
%   \includegraphics[width=1\linewidth]{image/corruptions/gaussian_obj.png}
%   \caption{ Point cloud under object-level \textit{gaussian} corruption and the \textit{Car} detection by PVRCNN ({\color{red}Red} BBoxes for the ground-truth and {\color{green}green} ones for the PVRCNN detection)}
%   \label{fig: corruptions_gaussian_obj}
% \end{figure}

% {
% {\noindent \bf \textit{Impulse\_obj}.} For \textit{impulse\_obj}, we add deterministic perturbations of $+/- 0.1m$ to the Cartesian coordinates of a certain portion of points of annotated objects in the point cloud. The portion  for 6 severity levels, where $N$ represents the number of points. Figure S21 displays the point cloud example under \textit{impulse\_obj}. 
% }

% \begin{figure}[h]
%   \centering
%   \includegraphics[width=1\linewidth]{image/corruptions/impulse_obj.png}
%   \caption{ Point cloud under object-level \textit{impulse} corruption and the \textit{Car} detection by PVRCNN ({\color{red}Red} BBoxes for the ground-truth and {\color{green}green} ones for the PVRCNN detection)}
%   \label{fig: corruptions_impulse_obj}
% \end{figure}

{
{\noindent \bf \textit{Upsample\_obj}.} We upsampled points (with the spatial bias within [-0.05, 0.05]) nearby a certain portion of points of annotated objects. The portion is set to $\{0, N/5, N/4, N/3, N/2, N\}$, where $N$ represents the number of original points of objects. Figure S22 displays the point cloud example under \textit{upsample\_obj}.
}

% \begin{figure}[h]
%   \centering
%   \includegraphics[width=1\linewidth]{image/corruptions/upsample_obj.png}
%   \caption{ Point cloud under object-level \textit{upsample} corruption and the \textit{Car} detection by PVRCNN ({\color{red}Red} BBoxes for the ground-truth and {\color{green}green} ones for the PVRCNN detection)}
%   \label{fig: corruptions_upsample_obj}
% \end{figure}

% \subsubsection{Object-level Density Corruption}

{
{\noindent \bf \textit{Cutout\_obj}.} For \textit{cutout\_obj}, we erase the distance-related neighborhoods of selected points from every annotated object in the point cloud. For the individual object, the number of selected points and the number of neighbor points are set to $\{(0, 0), (1,20), (2,20), (3,20), (4,20), (5,20)\}$. Figure S23 displays the point cloud example under \textit{cutout\_obj}.
}

% \begin{figure}[h]
%   \centering
%   \includegraphics[width=1\linewidth]{image/corruptions/cutout_obj.png}
%   \caption{ Point cloud under object-level \textit{cutout} corruption and the \textit{Car} detection by PVRCNN ({\color{red}Red} BBoxes for the ground-truth and {\color{green}green} ones for the PVRCNN detection)}
%   \label{fig: corruptions_cutout_obj}
% \end{figure}

{
{\noindent \bf \textit{Local\_dec\_obj} and \textit{Local\_inc\_obj}.} For \textit{local\_dec\_obj}, we randomly select a certain number of points of every annotated object as centers, and delete the $75\%$ points of the spatial neighborhood of every center. For \textit{local\_inc\_obj}, within the neighborhood of every centers on objects, we utilize linearly fitting to upsample points as many as the neighbor points. The number of selected points and the number of neighbor points are set to $\{(0, 0), (1,30), (2,30), (3,30), (4,30), (5,30)\}$ for \textit{local\_dec\_obj} and \textit{local\_inc\_obj}. Figures S24 and S25 display the point cloud examples under \textit{local\_dec\_obj} and \textit{local\_inc\_obj}.
}

% \begin{figure}[h]
%   \centering
%   \includegraphics[width=1\linewidth]{image/corruptions/local_dec_obj.png}
%   \caption{ Point cloud under object-level \textit{local\_dec} corruption and the \textit{Car} detection by PVRCNN ({\color{red}Red} BBoxes for the ground-truth and {\color{green}green} ones for the PVRCNN detection)}
%   \label{fig: corruptions_local_dec_obj}
% \end{figure}

% \begin{figure}[h]
%   \centering
%   \includegraphics[width=1\linewidth]{image/corruptions/local_inc_obj.png}
%   \caption{ Point cloud under object-level \textit{local\_inc} corruption and the \textit{Car} detection by PVRCNN ({\color{red}Red} BBoxes for the ground-truth and {\color{green}green} ones for the PVRCNN detection)}
%   \label{fig: corruptions_local_inc_obj}
% \end{figure}

{
{\noindent \bf \textit{Shear}.} For \textit{shear}, we slant points of objects on X and Y-axis by a transformation matrix $A = [[1, a, b], [c, 1, d], [0, 0, 1]]$ where $a, b, c, d$ are $+/-1$ $\times$ a float sampled on the uniform distribution. The upper and lower boundaries of the uniform distribution are set to \{(0, 0), (0, 0.10), (0.05, 0.15), (0.10, 0.20), (0.15, 0.25), (0.20, 0.30)\}. Figure S26 displays the point cloud example under \textit{shear}. 
}

% \begin{figure}[h]
%   \centering
%   \includegraphics[width=1\linewidth]{image/corruptions/shear.png}
%   \caption{ Point cloud under \textit{shear} corruption and the \textit{Car} detection by PVRCNN ({\color{red}Red} BBoxes for the ground-truth and {\color{green}green} ones for the PVRCNN detection)}
%   \label{fig: corruptions_shear}
% \end{figure}

{
{\noindent \bf \textit{FFD}.} We use the \textit{FFD} tool of \textit{pygem} to distort points of objects. With the prior setting of $5\times5\times5$ control points, the distortion ratio is sampled on the uniform distribution. The upper and lower boundaries of the uniform distribution are set to $+/- \{0, 0.1, 0.2, 0.3, 0.4, 0.5\}$. Figure S27 displays the point cloud example under \textit{FFD}. 
}

% \begin{figure}[h]
%   \centering
%   \includegraphics[width=1\linewidth]{image/corruptions/FFD.png}
%   \caption{ Point cloud under \textit{FFD} corruption and the \textit{Car} detection by PVRCNN ({\color{red}Red} BBoxes for the ground-truth and {\color{green}green} ones for the PVRCNN detection)}
%   \label{fig: corruptions_FFD}
% \end{figure}

{
{\noindent \bf \textit{Scale}.} Along the randomly selected axis (height, length, or width), we scale up or down points of objects by a transformation matrix $A = [[xs, 0, 0], [0, ys, 0], [0, 0, zs]]$. The scaling parameter randomly selected among $xs, ys, zs$ are set to $ 1 +/- \{0, 0.04, 0.08, 0.12, 0.16, 0.20\}$. Note that for scaling on Z-axis, we correspondingly move the object to the ground. Also, the ground-truth labels of objects (specifically, dimensions and locations of BBoxes) are modified accordingly. Figure S28 displays the point cloud example under \textit{scale}. 
}

% \begin{figure}[h]
%   \centering
%   \includegraphics[width=1\linewidth]{image/corruptions/scale.png}
%   \caption{ Point cloud under \textit{scale} corruption and the \textit{Car} detection by PVRCNN ({\color{red}Red} BBoxes for the ground-truth and {\color{green}green} ones for the PVRCNN detection)}
%   \label{fig: corruptions_scale}
% \end{figure}

{
{\noindent \bf \textit{Rotation} and \textit{Translation}.} We rotate or translate annotated objects to a milder degree. Specifically, objects are 1) moved forward or backward on the X and Y-axis at a distance sampled on the uniform distribution $U_{distance}$, and are 2)rotated in a clockwise or anticlockwise direction to a degree sampled on the uniform distribution $U_{degree}$. The lower and upper boundaries of $U_{degree}$ are set to \{(0, 0), (0, 2), (3, 4), (5, 6), (7, 8), (9, 10)\}degree and those of $U_{distance}$ to \{(0, 0), (0.0, 0.2), (0.3, 0.4), (0.5, 0.6), (0.7, 0.8), (0.9, 0.1)\}m.
Note that the ground-truth labels of objects are modified accordingly. Figures S29 and S30 display the examples under \textit{rotation} and \textit{translation}.
}

% \begin{figure}[h]
%   \centering
%   \includegraphics[width=1\linewidth]{image/corruptions/rotation.png}
%   \caption{ Point cloud under \textit{rotation} corruption and the \textit{Car} detection by PVRCNN ({\color{red}Red} BBoxes for the ground-truth and {\color{green}green} ones for the PVRCNN detection)}
%   \label{fig: corruptions_rotation}
% \end{figure}

% \begin{figure}[h]
%   \centering
%   \includegraphics[width=1\linewidth]{image/corruptions/translation.png}
%   \caption{ Point cloud under \textit{translation} corruption and the \textit{Car} detection by PVRCNN ({\color{red}Red} BBoxes for the ground-truth and {\color{green}green} ones for the PVRCNN detection)}
%   \label{fig: corruptions_translation}
% \end{figure}

\end{appendices}

\end{document}